\documentclass{article}
\usepackage{ijcai25}

\usepackage{times}
\usepackage{soul}
\usepackage{url}
\usepackage{hyperref}
\usepackage[utf8]{inputenc}
\usepackage[small]{caption}
\usepackage{graphicx}
\usepackage{amsmath}
\usepackage{amsthm}
\usepackage{booktabs}
\usepackage{algorithm}
\usepackage{algorithmic}
\usepackage[switch]{lineno}

\usepackage[T1]{fontenc}    
\usepackage{amsfonts}       
\usepackage{nicefrac}       
\usepackage{microtype}      
\usepackage{xcolor}         

\usepackage{amsthm}         
\usepackage{graphicx}       
\usepackage{multirow}
\usepackage{amsthm,amsmath}
\usepackage{mathrsfs}
\usepackage{algorithm}
\usepackage{algorithmic}
\usepackage{bm}
\usepackage{caption}
\usepackage{graphicx}
\usepackage{float} 
\usepackage{subcaption}
\usepackage{enumitem}
\usepackage{setspace}

\newtheorem{theorem}{Theorem}
\newtheorem{assumption}{Assumption}
\newtheorem{remark}{Remark}

\title{Fast Second-Order Online Kernel Learning Through\\ Incremental Matrix Sketching and Decomposition}

\author{
Dongxie Wen$^1$
\and
Xiao Zhang$^{1,}$\thanks{Xiao Zhang and Zhewei Wei are the corresponding authors.}\and
Zhewei Wei$^{1,*}$ \and
Chenping Hou$^2$\and
Shuai Li$^3$\And
Weinan Zhang$^3$\\
\affiliations
$^1$Gaoling School of Artificial Intelligence, Renmin University of China, Beijing, China\\
$^2$National University of Defense Technology\\
$^3$Shanghai Jiao Tong University\\
\emails
\{2019202221, zhangx89, zhewei\}@ruc.edu.cn,
hcpnudt@hotmail.com,
\{shuaili8, wnzhang\}@sjtu.edu.cn
}

\begin{document}

\maketitle

\begin{abstract}
    Second-order Online Kernel Learning (OKL) has attracted considerable research interest due to its promising predictive performance in streaming environments. However, existing second-order OKL approaches suffer from at least quadratic time complexity with respect to the pre-set budget, rendering them unsuitable for large-scale datasets.
    Moreover, the singular value decomposition required to obtain explicit feature mapping is computationally expensive due to the complete decomposition process.
    To address these issues, we propose FORKS, a fast incremental matrix sketching and decomposition approach tailored for second-order OKL.
    FORKS constructs an incremental maintenance paradigm for second-order kernelized gradient descent, which includes incremental matrix sketching for kernel approximation and incremental matrix decomposition for explicit feature mapping construction. Theoretical analysis demonstrates that FORKS achieves a logarithmic regret guarantee on par with other second-order approaches while maintaining a linear time complexity w.r.t. the budget, significantly enhancing efficiency over existing methods. We validate the performance of our method through extensive experiments conducted on real-world datasets, demonstrating its superior scalability and robustness against adversarial attacks.
\end{abstract}

\section{Introduction}

The objective of online learning is to efficiently and effectively update hypotheses in a data stream environment, where the processes of training and testing are intermixed \cite{Shalev2011OLA}. 
A popular online learning algorithm is Online Gradient Descent (OGD)~~\cite{zinkevich2003online}, which aims to minimize the loss function by iteratively adjusting the parameters in the direction of the negative gradient of the function. 
To address this limitation, Online Kernel Learning (OKL) maps the input space to a high-dimensional reproducing kernel Hilbert space (RKHS), effectively handling nonlinear learning tasks~\cite{kivinen2004online,Singh2012Online,Hu2015Kernelized,Lu2016Online,LeleCAO:276,Sahoo2019Large}.

From an optimization perspective, OKL can be classified into first-order and second-order methods. 
First-order methods use a fixed learning rate for gradient descent, resulting in \(O(\sqrt{T})\) regret for any arbitrary sequence of convex losses~\cite{cavallanti2007tracking,orabona2008projectron,zhao2012fast,Lu2016Online,zhang2019incremental}, where $T$ denotes the number of rounds. 
Although some works set a more aggressive learning rate to improve the regret bound to $O(\log T)$, it requires the assumption that the loss function exhibits strong convexity, which is unrealistic for most loss functions~\cite{zhu2015online}.
In contrast, second-order algorithms utilize the Hessian matrix to adjust the learning rate dynamically, achieving logarithmic regret without requiring strong convexity in all directions~\cite{hazan2007logarithmic,zhdanov2013identity,calandriello2017efficient,calandriello2017second,zhang2023reward}.

Despite the improved regret, exact second-order OKL requires \(O(t^2)\) space and time complexity at round $t$ due to the need to store the entire kernel matrix. 
Previous studies have primarily focused on approximating the kernel matrix using online sampling techniques~\cite{calandriello2017efficient,calandriello2017second,ChengCHEN:165330}. 
However, sampling-based algorithms face two significant issues.
First, without prior knowledge of the effective dimension of streaming data, sampling-based methods exhibit at least quadratic time complexity with respect to the budget, which is the maximum size of the static subspace. 
Second, these methods only address the problem of kernel matrix approximation in online kernel learning and cannot efficiently obtain explicit feature mapping. More precisely, due to the varying size of the subspace, complete singular value decomposition is required to calculate the feature mapping, which is computationally expensive.

In this paper, we propose a fast incremental matrix sketching approach for second-order online kernel learning, along with an efficient decomposition method designed for incremental updates, effectively addressing the two aforementioned challenges both theoretically and experimentally.
Our contributions can be summarized as follows:
\begin{itemize}
    \item We propose FORKS, a fast and effective second-order online kernel learning method that can be generalized to both regression and classification tasks. 
    FORKS maintains incremental matrix sketching using efficient low-rank modifications and constructs an effective time-varying explicit feature mapping. 
    We provide a detailed theoretical analysis to illustrate the advantages of FORKS, including having linear time complexity w.r.t. the budget, and enjoying a logarithmic regret bound.
    \item We propose TISVD, a novel incremental singular value decomposition adapting to matrix decomposition problems in online learning environments. 
    We theoretically compare the time complexity between TISVD and the original truncated low-rank SVD, confirming that FORKS with TISVD is computationally more efficient without compromising prediction performance.
    \item We conduct extensive experiments to demonstrate the superior performance of FORKS on both adversarial and real-world datasets, while also enhancing efficiency. Furthermore, we validate FORKS's robustness and scalability using a large-scale streaming recommendation dataset.
\end{itemize}

\section{Notations and Preliminaries}

Let $[n] = \{1, 2, \ldots, n\}$, upper-case bold letters (e.g., $\bm{A}$) represent matrix and lower-case bold letters (e.g., $\bm{a}$) represent vectors.
We denote by $\bm{A}_{i*}$ and $\bm{A}_{*j}$ the $i$-th row and $j$-th column of matrix $\bm{A}$, $\bm{A}^{\dagger}$ the Moore-Penrose pseudoinverse of $\bm{A}$, $\bm{\Vert A\Vert}_2$ and $\bm{\Vert A\Vert}_F$ the spectral and Frobenius norms of $\bm{A}$. 
Let $\mathcal{S} = \{(\bm{x}_t,y_t)\}_{t=1}^T \subseteq (\mathcal{X} \times \mathcal{Y})^T$ be the data stream of $T$ instances, where $\bm{x}_t\in \mathbb{R}^d$. 
We use $\bm{A=U\Sigma{V^{\top}}}$ to represent the SVD of $\bm{A}$, where $\bm{U}, \bm {V}$ denote the left and right matrices of singular vectors and $\bm{\Sigma} = \text{diag}[\lambda_1,...,\lambda_n]$ is the diagonal matrix of singular values. 

\subsection{Online Kernel Learning}
In this section, we introduce the problem of online kernel learning.
Let the kernel function be denoted by \(\kappa : \mathcal{X} \times \mathcal{X} \to \mathbb{R}\), with the corresponding kernel matrix \(\bm{K} = \left( \kappa(\bm{x}_i, \bm{x}_j) \right)\), where \(\mathcal{X}\) represents the input space. Let \(\mathcal{H}_{\kappa}\) denote the reproducing kernel Hilbert space (RKHS) induced by \(\kappa\), and let the feature mapping \(\boldsymbol{\phi} : \mathcal{X} \to \mathcal{H}_{\kappa}\) correspond to this RKHS. In this context, the kernel function can be expressed as the inner product \(\kappa(\bm{x}_i, \bm{x}_j) = \boldsymbol{\phi}(\bm{x}_i)^{\top} \boldsymbol{\phi}(\bm{x}_j)\).

Consider a data stream \(\mathcal{S}\) and a convex loss function \(\ell\). At round \(t\), we define the hypothesis as \(f_t = \bm{w}_t^{\top} \boldsymbol{\phi}(\bm{x}_t)\), where \(\bm{x}_t\) is the incoming sample. Upon receiving a new example \(\bm{x}_t\), the hypothesis predicts the label \(\hat{y_t}\) using \(f_t\). The hypothesis incurs a loss \(\ell_t(f_t(\bm{x}_t)) := \ell(f_t(\bm{x}_t), y_t)\), where \(y_t\) is the true label, and subsequently updates its model parameters. 
The objective of an online learning algorithm is to bound the cumulative regret, which is defined as: 
\[
\mathrm{Reg}_T(f^*) = \sum_{t=1}^T \left[ \ell_t(f_t) - \ell_t(f^*) \right],
\]
where \(f^*\) is the optimal hypothesis, determined in hindsight.

\subsection{Matrix Sketching}
\label{sec:matrix sketching}
Given a matrix $\bm{M}\in\mathbb{R}^{a\times{b}}$, the sketch of $\bm{M}$ is defined as $\bm{M}\bm{S}\in{\mathbb{R}^{a\times{s}}}$, where $\bm{S}\in\mathbb{R}^{b\times{s}}$ is a sketch matrix.
In this paper, we introduce the Sparse Johnson-Lindenstrauss Transform and Column-sampling matrix as the sketch matrix~\cite{charikar2002finding,kane2014sparser}.

\paragraph{Sparse Johnson-Lindenstrauss Transform (SJLT).}
SJLT is a randomized sketching technique based on hash functions. 
SJLT consists of \(D\) submatrices, written as \(\bm{S} = \left[\bm{S}_1, \dots, \bm{S}_D\right] \in \mathbb{R}^{b \times s_p}\). Each submatrix $\bm{S}_k\in\mathbb{R}^{b\times\left(s_p/D\right)}$ is defined by two sets of hash functions:

\begin{align*}
    &h_k: \{1, \dots, b\} \rightarrow \left\{1, \dots, \frac{s_p}{D}\right\} ,\\
    &g_k: \{1, \dots, b\} \rightarrow \left\{\frac{-1}{\sqrt{D}},\frac{1}{\sqrt{D}}\right\},
\end{align*}
where \(k \in \{1, \dots, D\}\).
For each submatrix, the element \([\bm{S}_k]_{i,j}\) equals \(g_k(i)\) if \(j = h_k(i)\), and equals 0 otherwise.

\paragraph{Column-sampling matrix.}
We denote the column-sampling matrix by $\bm{S}_m\in\mathbb{R}^{b\times{s_m}}$, the columns of $\bm{S}_m$ is obtained by uniformly sampling column vectors of $\bm{I}_{b\times{b}}$.

\section{FORKS: The Proposed Algorithm}
\label{sec: FORKS: The Proposed Algorithm}
In this section, we propose a novel and efficient second-order OKL method that incorporates incremental sketching and decomposition. While previous work has primarily focused on the incremental maintenance of the approximated kernel matrix, our approach is the first to address the incremental construction and maintenance of the feature mapping.

\subsection{Algorithm Overview}
\begin{figure}[htp]
\centering
\includegraphics[width=.47\textwidth]{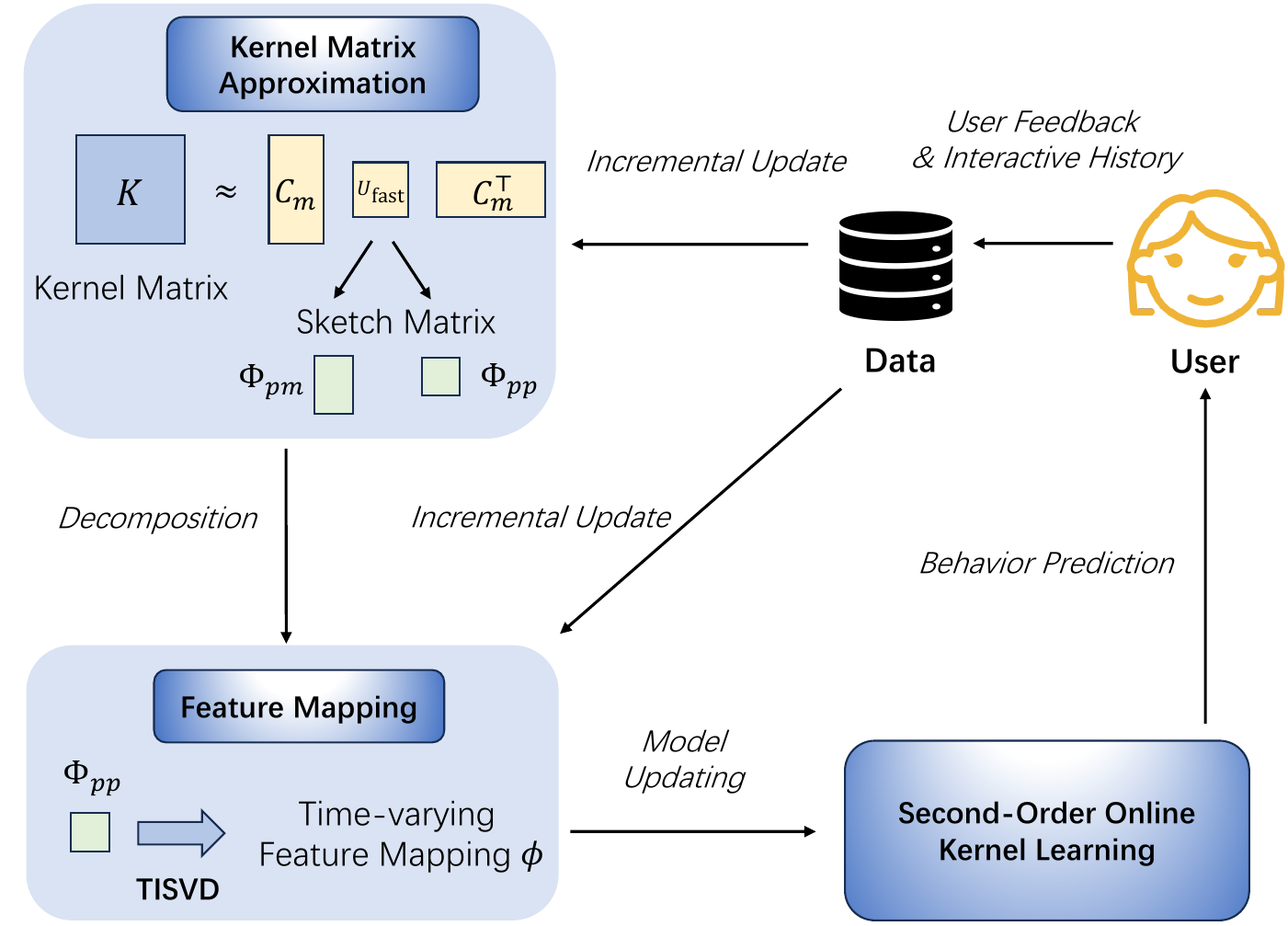}
 \caption{The illustration of the proposed method FORKS.
 }  
 \label{fig: framework}
\end{figure}

The second-order OKL process can be described in three key stages: kernel matrix approximation, feature mapping, and second-order online learning. Figure~\ref{fig: framework} illustrates the overall architecture of our method.
Rather than storing the entire kernel matrix, we employ matrix sketching techniques to approximate it with a fixed, constant size.
Furthermore, the novel truncated incremental singular value decomposition method is utilized to generate the time-varying feature mapping.
Notably, both the kernel matrix and feature mapping can be updated incrementally, ensuring efficient maintenance and adaptability to changes in user preferences. Additionally, we use second-order updates to predict user behavior effectively.
A non-trivial proof establishes that our method guarantees logarithmic regret while preserving computational efficiency.

\subsection{Matrix Sketching for Kernel Matrix Approximation}
In this section, we apply the matrix sketching technique to approximate the kernel matrix.
Although this reduction has been employed in first-order methods~\cite{Zhang2018Online,zhang2019incremental}, the regret bound for second-order online kernel learning, which operates under the directional curvature condition, remains unknown in the literature.

In the offline setting, given the kernel matrix \(\bm{K}^{(t)} \in \mathbb{R}^{t \times t}\), the prototype model~\cite{williams2000using} computes the approximate kernel matrix \(\hat{\bm{K}}^{(t)} = \bm{C}_m^{(t)} \bm{U}_\text{fast} (\bm{C}_m^{(t)})^{\top}\) by solving the following optimization problem at round \(t\):
\begin{equation}
\label{prototype}
\begin{aligned}
    \bm{U}_\text{fast} &= 
    \operatornamewithlimits{arg\,min}\limits_{\bm U}\left\Vert\bm{C}_m^{(t)}\bm{U}\left(\bm{C}_m^{(t)}\right)^{\top}-\bm{K}^{(t)}\right\Vert_{F}^2,
\end{aligned}
\end{equation}
where \(\bm{C}_m^{(t)}\) represents the sketch matrix at round $t$, defined as \(\bm{C}_m^{(t)} = \bm{K}^{(t)} \bm{S}_m^{(t)} \in \mathbb{R}^{t \times s_m}\), with \(\bm{S}_m^{(t)} \in \mathbb{R}^{t \times s_m}\) being the column-sampling matrix used to reduce the size of the approximate kernel matrix.

Note that solving Eq.~\eqref{prototype} can impose substantial computational demands. To address this problem, the sketch-based method~\cite{wang2016spsd} is proposed to reduce the computational complexity. 
Specifically, let \(\bm{S}_p^{(t)} \in \mathbb{R}^{t \times s_p}\) represent the randomized sketch matrix based on the SJLT (details in Section \ref{sec:matrix sketching}).
The sketched kernel matrix approximation problem at round $t$ is then formulated as follows:
\begin{equation}
\begin{aligned}
    \label{KernelApproximation}
    \bm{U}_\text{fast}
    &= 
    \operatornamewithlimits{arg\,min}\limits_{\bm U}\Big\Vert\left(\bm{S}_p^{(t)}\right)^{\top}\bm{C}_m^{(t)}\bm{U}\left(\bm{C}_m^{(t)}\right)^{\top}\bm{S}_p^{(t)}- \\
    &\hphantom{{}={}}
    \left(\bm{S}_p^{(t)}\right)^{\top}\bm{K}^{(t)}\bm{S}_p^{(t)}\Big\Vert_{F}^2 \\
    &=\left(\boldsymbol{\Phi}_{pm}^{(t)}\right)^{\dagger}\boldsymbol{\Phi}_{pp}^{(t)}\left(\boldsymbol{\Phi}_{pm}^{(t)\top}\right)^{\dagger}
    ,
\end{aligned}
\end{equation}
where
\begin{equation}
\label{InitSketch}
    \begin{aligned}
         \boldsymbol{\Phi}_{pm}^{(t)} &= \bm{S}_p^{(t)\top}\bm{C}_m^{(t)}\in\mathbb{R}^{s_p\times{s_m}},\\\boldsymbol{\Phi}_{pp}^{(t)} &= \bm{S}_p^{(t)\top}\bm{K}^{(t)}\bm{S}_p^{(t)}\in\mathbb{R}^{s_p\times{s_p}}.
    \end{aligned}
\end{equation}

Instead of storing the entire kernel matrix, we can maintain smaller sketches for approximation. 
In the online setting, these sketches can be incrementally updated with the arrival of new data, \(\bm{x}_{t+1}\). Specifically, the updates are given by:
\begin{equation}
\label{eq: sketch update}
    \begin{aligned}
    \boldsymbol{\Phi}_{pm}^{(t+1)}&=\boldsymbol{\Phi}_{pm}^{(t)}+\bm{\Delta}_{pm}^{(t+1)},\\
    \boldsymbol{\Phi}_{pp}^{(t+1)}&=\boldsymbol{\Phi}_{pp}^{(t)}+\bm{\Delta}_{pp}^{(t+1)},
    \end{aligned}
\end{equation}
where \(\bm{\Delta}_{pm}^{(t+1)}\) and \(\bm{\Delta}_{pp}^{(t+1)}\) are composed of three rank-1 matrices. 
Due to space constraints, the detailed construction is provided in Appendix B.1.

Next, we construct the time-varying explicit feature mapping.
For simplicity, we use rank-\(k\) SVD, though it is inefficient due to the need for a full matrix decomposition.
The elements of the kernel matrix are equal to the inner product of the corresponding points after feature mapping, i.e. $\bm{K}_{i,j} = \bm{\phi}({x_i})^{\top}\bm{\phi}({x_j})$.
Once we build the approximate kernel matrix by Eq.~\eqref{KernelApproximation}, we can obtain a time-varying feature mapping through the rank-$k$ SVD. 
Specifically, if
\begin{equation}
\label{Qsvd}
    \boldsymbol{\Phi}_{pp}^{(t+1)} \approx \bm{V}^{(t+1)}\boldsymbol{\Sigma}^{(t+1)}\bm{V}^{(t+1)\top} \in\mathbb{R}^{s_p\times{s_p}},
\end{equation}
where $\bm{V}^{(t+1)} \in \mathbb{R}^{s_p\times{k}}, \boldsymbol{\Sigma}^{(t+1)} \in \mathbb{R}^{k\times{k}}$ and rank $k\le{s_p}$, 
we can update the time-varying explicit feature mapping $\boldsymbol{\phi}_{t+2}\in\mathbb{R}^{k}$ at round $t+1$ by
\begin{equation*}
    \boldsymbol{\phi}_{t+2}(\cdot) = \left([\kappa(\cdot,\tilde{\bm{x}}_1),\kappa(\cdot,\tilde{\bm{x}}_2),...,\kappa(\cdot,\tilde{\bm{x}}_{s_m})]\bm{Z}_{t+1}\right)^{\top} ,
\end{equation*}
where $\{\tilde{\bm{x}}_{i}\}_{i=1}^{s_m}$ are the sampled columns by $\bm{S}_m^{(t+1)}$ and
$
    \bm{Z}_{t+1} = (\boldsymbol{\Phi}_{pm}^{(t+1)})^{\dagger}\bm{V}^{(t+1)}(\boldsymbol{\Sigma}^{(t+1)})^{\frac{1}{2}}.
$

However, directly applying rank-\(k\) SVD to \(\boldsymbol{\Phi}_{pp}\) is inefficient in online learning scenarios. Specifically, the standard rank-\(k\) SVD incurs a time complexity of \(O(s_p^3)\) at each update, rendering it impractical in scenarios with frequent updates.

\subsection{Novel Incremental Matrix Decomposition Method for Feature Mapping}

In this section, we propose \textbf{TISVD} (\underline{T}runcated \underline{I}ncremental \underline{S}ingular \underline{V}alue \underline{D}ecomposition), a novel incremental SVD method designed for decomposing sketches. 
TISVD achieves linear time and space complexity with respect to the sketch size \(s_p\), effectively avoiding the time-consuming operation of performing a complete SVD.

We begin by presenting the construction of TISVD, which is well-suited for decomposing matrices with low-rank update properties.
Without loss of generality, we denote a matrix at round $t$ as $\bm{M}^{(t)} = \bm{U}^{(t)}\boldsymbol{\Sigma}^{(t)}\bm{V}^{(t)\top}$.
In the $(t+1)$-th round, $\bm{M}^{(t)}$ is updated by low-rank matrices as follows:
\begin{equation}
\label{TISVD_matrix_pp}
\begin{aligned}
    \bm{M}^{(t+1)} &= \bm{M}^{(t)} + \bm{\Delta}_1\bm{\Delta}_2^{\top} \\&= 
    \bm{U}^{(t+1)}\boldsymbol{\Sigma}^{(t+1)}\left(\bm{V}^{(t+1)}\right)^{\top},
\end{aligned}
\end{equation}
where $\bm{\Delta}_1,\bm{\Delta}_2 \in \mathbb{R}^{s_p\times{c}}$ of rank $r\le{c}\ll{s_p}$.

Our objective is to directly update the singular matrices $\bm{U}^{(t)}$,  $\boldsymbol{\Sigma}^{(t)}$ and $\bm{V}^{(t)}$ using low-rank update matrices $\bm{\Delta}_1$ and $\bm{\Delta}_2$, resulting in $\bm{U}^{(t+1)}$, $\boldsymbol{\Sigma}^{(t+1)}$ and $\bm{V}^{(t+1)}$.
First, we formulate orthogonal matrices through orthogonal projection and vertical projection.
Let $\bm{P}, \bm{Q}$ denote the orthogonal basis of the column space of the following matrices:
$$\Big(\bm{I}-\bm{U}^{(t)}\bm{U}^{(t)\top}\Big)\bm{\Delta}_1
\quad \text{and} \quad \Big(\bm{I}-\bm{V}^{(t)}\bm{V}^{(t)\top}\Big)\bm{\Delta}_2,
$$ respectively.

Set 
$\bm{R}_1 \doteq \bm{P}^{\top}\Big(\bm{I}-\bm{U}^{(t)}\bm{U}^{(t)\top}\Big)\bm{\Delta}_1$ and $\bm{R}_2 \doteq \bm{Q}^{\top}\Big(\bm{I}-\bm{V}^{(t)}\bm{V}^{(t)\top}\Big)\bm{\Delta}_2$,
we can transform Eq.~\eqref{TISVD_matrix_pp} into
\begin{equation}
    \label{ISVD_QR}
    \bm{M}^{(t+1)} =
    \begin{bmatrix}
    \bm{U}^{(t)} & \bm{P} 
    \end{bmatrix}
    \bm{H}
    \begin{bmatrix}
    \bm{V}^{(t)} & \bm{Q} 
    \end{bmatrix}^{\top},
\end{equation}
where
\begin{equation}
    \label{TISVD_H}
    \bm{H} = \begin{bmatrix}
    \boldsymbol{\Sigma}^{(t)}& \bm{0} \\
    \bm{0} & \bm{0}
    \end{bmatrix} + 
    \begin{bmatrix}
    \bm{U}^{(t)\top}\bm{\Delta}_1 \\
    \bm{R}_1
    \end{bmatrix}
    \begin{bmatrix}
    \bm{V}^{(t)\top}\bm{\Delta}_2 \\
    \bm{R}_2
    \end{bmatrix}^{\top} .
\end{equation}

Subsequently, as the size of $\bm{H}$ is smaller than $\bm{M}^{(t+1)}$, an efficient computation of $\tilde{\bm{U}}_k$, $\tilde{\bm{V}}_k$, and $\tilde{\boldsymbol{\Sigma}}_k$ can be obtained by performing a truncated rank-$k$ SVD on $\bm{H}$. 
Since the matrices on the left and right sides are column orthogonal, we finally obtain $\bm{U}^{(t+1)}$, $\bm{V}^{(t+1)}$, and $\boldsymbol{\Sigma}^{(t+1)}$ at round $t+1$:
\begin{equation}
\label{TISVD_UVS}
    \begin{aligned}
    \bm{U}^{(t+1)} &= \begin{bmatrix}
    \bm{U}^{(t)} & \bm{P} 
    \end{bmatrix}\tilde{\bm{U}}_k,
    \\
    \bm{V}^{(t+1)} &= \begin{bmatrix}
    \bm{V}^{(t)} & \bm{Q} 
    \end{bmatrix}\tilde{\bm{V}}_k,\\
    \boldsymbol{\Sigma}^{(t+1)} &= \tilde{\boldsymbol{\Sigma}}_k.
\end{aligned}
\end{equation}

We summarize the algorithm and provide the pseudo-code for TISVD in Appendix B.2. 
Note that the sketch \(\boldsymbol{\Phi}_{pp}^{(t+1)}\) can be updated using low-rank matrices, as shown in Eq.~\eqref{eq: sketch update}. By replacing the standard rank-\(k\) SVD with TISVD, we establish an efficient mechanism for the incremental maintenance of singular matrices. More precisely, in Eq.~\eqref{Qsvd}, we update \(\bm{V}^{(t+1)}\) and \(\boldsymbol{\Sigma}^{(t+1)}\) using their previous counterparts, \(\bm{V}^{(t)}\) and \(\boldsymbol{\Sigma}^{(t)}\), along with the low-rank update \(\bm{\Delta}_{pp}^{(t+1)}\) at round \(t+1\).

\begin{remark}[Complexity]
    The complexity of TISVD is primarily composed of the truncated rank-\(k\) SVD on \(\bm{H}\) and the matrix multiplication in Eq.~\eqref{TISVD_UVS}. 
    The update cost of the truncated rank-\(k\) SVD is \(O(k^3)\), while the matrix multiplication in Eq.~\eqref{TISVD_UVS} incurs a complexity of \(O(s_p k)\). 
    Consequently, compared to standard rank-\(k\) SVD, TISVD provides significant improvements by reducing the time complexity from \(O(s_p^3)\) to \(O(s_p k + k^3)\) and the space complexity from \(O(s_p^2)\) to \(O(s_p k + k^2)\). 
    More discussion is provided in Appendix B.3.
\end{remark}

\subsection{Efficient Second-Order Online Kernel Learning}
Since the efficient time-varying explicit feature mapping $\boldsymbol{\phi}_t(\cdot)$ has been constructed, we can formulate the approximate hypothesis $f_t(\bm{x}_t)$ at round $t$ as follows
\begin{equation*}
    f_t(\bm{x}_t) = \bm{w}_t^{\top}\boldsymbol{\phi}_t(\bm{x}_t) ,
\end{equation*} 
where $\bm{w}_t$ is the weight vector.
On the basis of the hypothesis, we propose a two-stage second-order online kernel learning method, named \textbf{FORKS} (\underline{F}ast Second-\underline{Or}der Online \underline{K}ernel Learning Using Incremental \underline{S}ketching).

In the first stage, we simply collect the items with nonzero losses to the buffer SV and perform the Kernelized Online Gradient Descent (KOGD)~\cite{kivinen2004online}.
When the size of the buffer reaches a fixed budget $B$, we calculate $\bm{K}_t$ and initialize sketch matrices $\boldsymbol{\Phi}_{pp}^{(t)}, \boldsymbol{\Phi}_{pm}^{(t)}$ in Eq.~\eqref{InitSketch}.

In the second stage, we adopt a periodic updating strategy for sketches.
More precisely, we update $\boldsymbol{\Phi}_{pp}^{(t)}$, $\boldsymbol{\Phi}_{pm}^{(t)}$ by Eq.~\eqref{eq: sketch update} once for every $\rho$ examples, where $\rho\in[T-B]$ is defined as update cycle.
Furthermore, we incrementally update the feature mapping $\boldsymbol{\phi}_t(\cdot)$ by TISVD.

In addition to updating the feature mapping $\boldsymbol{\phi}_t(\cdot)$, we perform second-order updates on the $\bm{w}_t$.
Specifically, we update the hypothesis using Online Newton Step (ONS)~\cite{hazan2007logarithmic} for some parameters $\alpha>0$ and $\sigma_i,\eta_i\ge0$:
\begin{equation}
    \label{eq:ONS}
    \begin{aligned}
    \bm{v}_{t+1} &= \bm{w}_{t}-\bm{A}_{t}^{-1}\bm{g}_{t}
    ,\\
    \bm{w}_{t+1} &= \bm{v}_{t+1}-\frac{h\left(\boldsymbol{\phi}_{t+1}^{\top}\bm{v}_{t+1}\right)}{\boldsymbol{\phi}_{t+1}^{\top}\bm{A}_{t}^{-1}\boldsymbol{\phi}_{t+1}}\bm{A}_{t}^{-1}\boldsymbol{\phi}_{t+1} ,
    \end{aligned}
\end{equation}
where $\bm{g}_t = \nabla_{\bm{w}_t}\ell_t(\hat{y}_t)$, $\bm{A}_{t} = \alpha\bm{I}+\sum_{i=0}^t(\sigma_i+\eta_i){\bm{g}_i\bm{g}_i^{\top}}$ and $h(z) = \text{sign}(z)\max(|z|-C,0)$.
The second-order updates not only consider the gradient information but also utilize the curvature information of the loss function, leading to faster convergence rates.

At the start of a new update epoch, we incorporate a \emph{reset} step before applying the gradient descent in the new embedded space. 
We update the feature mapping $\boldsymbol{\phi}_t$ but reset $\bm{A}_t$ and $\bm{w}_t$.
This step is taken to ensure that our starting point cannot be influenced by the adversary. 
By leveraging efficient second-order updates, we can effectively converge to the optimal hypothesis within the current subspace. Furthermore, the reset of the descent procedure when transitioning between subspaces ensures a stable starting point and maintains a bounded regret throughout the entire process.
We summarize the above stages into Algorithm~\ref{alg: FORKS}.

The incremental update of the feature mapping, combined with the reset mechanism, ensures that we can effectively capture changes in user preferences over time. 
More specifically, our algorithm updates the feature vector to reflect the user's current state, while resetting the Hessian matrix mitigates the influence of outdated data on the learner.

\begin{algorithm}[tb]
    \caption{{\sf FORKS}}
    \label{alg: FORKS}
    \textbf{Input}: Data stream $\{(\bm{x}_t,y_t)\}_{t=1}^T$, sketch size $s_p$, sample size $s_m$, rank $k$, budget $B$, update cycle $\rho$,  regularizer $\alpha$ \\
    \textbf{Output}: Predicted label $\{\hat{y}_t\}_{t=1}^T$
    \begin{algorithmic}[1] 
    \FOR{$t \leftarrow 1, \ldots, T$}{
           \STATE Receive $\bm{x}_t$
           \WHILE{SV is not full}
           \STATE Update hypothesis by KOGD
           \STATE Add $\bm{x}_t$ to SV whenever the loss is nonzero
           \ENDWHILE
            \IF{SV is full}
            \STATE Initialize $\boldsymbol{\Phi}_{pp}^{(t)},\boldsymbol{\Phi}_{pm}^{(t)}$ as in Eq.~\eqref{InitSketch}
            \STATE Compute the mapping $\boldsymbol{\phi}_{t+1}$
            \ELSIF{ in the update round}
            \STATE Update $\boldsymbol{\phi}_{t+1}$ by TISVD (Algorithm 2)
            \STATE Set $\bm{A}_{t}\gets\alpha\bm{I}$,\quad $\bm{w}_{t}\gets\bm{0}$
            \ELSE
            \STATE $\boldsymbol{\Phi}_{pp}^{(t)}\gets\boldsymbol{\Phi}_{pp}^{(t-1)},\boldsymbol{\Phi}_{pm}^{(t)}\gets\boldsymbol{\Phi}_{pm}^{(t-1)}, \boldsymbol{\phi}_{t+1}\gets \boldsymbol{\phi}_{t}$
            \ENDIF
            \STATE Predict $\hat{y}_t = \text{sgn}\left(\boldsymbol{\phi}_t^{\top}\bm{w}_t\right)$\;
           \STATE \# Execute a second-order gradient descent 
           \STATE Obtain $\bm{g}_t \gets \nabla_{\bm w_t}\ell_t\left(\hat{y}_t\right)$
           \STATE Update $\bm{A}_{t+1} \gets \bm{A}_{t}+(\sigma_i+\eta_i)\bm{g}_t\bm{g}_t^{\top}$
           \STATE Compute $\bm{w}_{t+1}$, $\bm{v}_{t+1}$ using Eq.~\eqref{eq:ONS}
    }
    \ENDFOR
    \end{algorithmic}
\end{algorithm}

\subsection{Complexity Analysis of FORKS}
Given the budget of $B$, FORKS consists of three parts: (1) the first stage using KOGD, (2) the updating round in the second stage, and (3) the regular round in the second stage. 
At the first stage $(|SV| \le B)$, FORKS has constant time $O(B)$ and space complexities $O(B)$ per round.

The main computational complexity of FORKS during the update round stems from the matrix decomposition and inversion procedures.
These processes are necessary for updating the feature mapping and performing second-order updates, respectively.
TISVD reduces the time complexity of $\boldsymbol{\Phi}_{pp}$ decomposition from $O(s_p^3)$ to $O(s_pk+k^3)$, where $s_p$ is the sketch size of $\bm{S}_p$ and $k$ is the rank in TISVD.
A naive implementation of the second-order update requires $O(k^3)$ per-step time and has a space complexity of $O(k^2)$ necessary to store the Hessian $\bm{A}_t$. 
By taking advantage of the fact that $\bm{A}_t$ is rank-$1$ updated, we can reduce the per-step cost to $O(k^2)$.

We denote the update cycle as $\rho$ and $\mu = B+\lfloor{({T-B})/{\rho}}\rfloor$.
To summarize, the time complexity of FORKS at each updating round is 
\begin{equation*}
    O\left(\mu+s_pk^2+s_ms_pk+k^3\right),
\end{equation*} and the space complexity is $O(\mu+s_pk+s_ms_p+k^2)$, where $s_m$ is the sketch size of $\bm S_m$.

The primary time consumption for the online learning algorithm occurs during regular rounds. 
At each regular round, the time complexity of FORKS is \( O(s_m k + k^2) \). 
Given that \( s_m < s_p < B \), our algorithm achieves a time complexity of \( O(B k + k^2) \) per step. 
The current state-of-the-art second-order online kernel learning method, PROS-N-KONS, presents a time complexity of \( O(B^2) \) per step in its budget version \cite{calandriello2017efficient}. 
Moreover, due to the changing size of the subspace from online sampling, PROS-N-KONS must perform a complete SVD at each step. 
In contrast, FORKS introduces substantial advancements by reducing the time complexity from \( O(B^2) \) to \( O(B k + k^2) \).

\section{Regret Analysis}
\label{sec: Regret Analysis}
In this section, we provide the regret analysis for the proposed second-order online kernel learning algorithm. 
We begin by making the following assumptions about the loss functions.

\begin{assumption}[Lipschitz Continuity]
\label{assm:OKL:Lip}
$\ell$ is Lipschitz continuous with the Lipschitz constant $L_{\mathrm{Lip}}$, i.e.,
$\left\| \nabla \ell( \bm w) \right\|_2 \leq L_{\mathrm{Lip}}$.
\end{assumption}

\begin{assumption}[Directional Curvature]
\label{assm:OKL:Curvature}
Let $L_{\mathrm{Cur}} \geq 0$. Then, for any vectors $\bm w_1, \bm w_2$ and the convex function $\ell$, denote that $\Delta = \ell (\bm w_1)-\ell (\bm w_2)$, we have
\begin{equation*}
    \Delta
    \geq  \left\langle \nabla \ell( \bm w_2), \bm w_1 - \bm w_2 \right\rangle+
    \frac{L_{\mathrm{Cur}}}{2}
    \left\langle \nabla \ell( \bm w_2), \bm w_1 - \bm w_2 \right\rangle^2.
\end{equation*}
\end{assumption}

In practical scenarios, the assumption of \emph{strong convexity} may not always hold as it imposes constraints on the convexity of losses in all directions.
A more feasible approach is to relax this assumption by demanding strong convexity only in the gradient direction, which is a weaker condition as indicated by the two assumptions above. 
For example, exp-concave losses like squared loss and squared hinge loss satisfy Assumption~\ref{assm:OKL:Curvature}.
\begin{assumption}[Matrix Product Preserving]\label{assum:FORKS:Matrix}
    Let $\bm S_{p} \in \mathbb{R}^{T \times s_p}$ be a sketch matrix, 
    $\bm U_{\mathrm{m}} \in \mathbb{R}^{T \times s_{\mathrm{m}}}$ be a matrix with orthonormal columns,
    $\bm U_{\mathrm{m}}^\bot \in \mathbb{R}^{T \times (T - s_{\mathrm{m}})}$ be another matrix satisfying
    $ \bm U_{\mathrm{m}} \bm U_{\mathrm{m}}^\top + \bm U_{\mathrm{m}}^\bot (\bm U_{\mathrm{m}}^\bot)^\top = \bm I_T $ and
    $\bm U_{\mathrm{m}}^\top \bm U_{\mathrm{m}}^\bot = \bm O$,
and $\delta_i$ ($i =1, 2$) be the failure probabilities
        defined as follows: 
    \begin{equation*}
        \mathrm{Pr}\left\{
              \frac{\| \bm B_i \bm A_i - \bm B_i {\bm S_p} {\bm S_p}^{\top} \bm A_i   \|_{F}^2}{2\Vert\bm{B}_i\Vert_F^2\Vert\bm{A}_i\Vert_F^2}
            > 
            \frac{1}{\delta_i s_p}
            \right\}
            \leq \delta_i,~~i = 1, 2,
    \end{equation*}
    where 
    $\bm A_1 = \bm U_{\mathrm{m}}$,
    $\bm B_1 = \bm I_{T}$,
    $\bm A_2 = {\bm U_{\mathrm{m}}^{\bot}} {({\bm U_{\mathrm{m}}}^{\bot})}^\top \bm K$,
    $\bm B_2 = \bm U_{\mathrm{m}}^\top$, $\bm K \in \mathbb{R}^{T \times T}$ is a kernel matrix.
\end{assumption}
The conditions stated in Assumption~\ref{assum:FORKS:Matrix} can be satisfied by SJLT matrix~\cite{Woodruff2014SAA}. 
Given the loss $\ell_t (\bm w_t) := \ell_t (f_t) = \ell (f_t (\bm x_t), y_t), \forall t \in [T]$ satisfies Assumption~\ref{assm:OKL:Lip} and Assumption~\ref{assm:OKL:Curvature}, we bound the following regret:
\[
    \mathrm{Reg}_T (f^*) = \sum_{t=1}^T \left[ \ell_t (\bm w_t) - \ell_t (f^*) \right],
\]
where $f^* = \operatornamewithlimits{arg\,min}_{f \in \mathcal{H}_\kappa} \sum_{t=1}^{T} \ell_t (f)$ denotes the optimal hypothesis in hindsight in the original RKHS.

Let \(\bm{K} \in \mathbb{R}^{T \times T}\) be a kernel matrix with \(\kappa(\bm{x}_i, \bm{x}_j) \leq 1\) for all \(i, j\). 
Let \(\rho = \lfloor \theta (T - B) \rfloor\) denote the update cycle, where \(\theta \in (0, 1)\), and let \(k\) represent the rank in TISVD. 
Additionally, define \(C_{\mathrm{Coh}}\) as the coherence of the intersection matrix of \(\bm{K}\), which is constructed using \(B + \lfloor (T - B) / \rho \rfloor\) examples. Let \(\bm{v}_i\) be the \(i\)-th singular vector of \(\bm{K}\), the coherence is given by:
\[
    C_{\mathrm{Coh}} = \left( \frac{B + 1/\theta}{\text{rank}(\bm{K})} \right) \max_{i} \|\bm{v}_i\|_2^2.
\]
Note that $C_{\mathrm{Coh}}$ is independent of $T$ when $\rho = \lfloor \theta (T - B) \rfloor$.
We demonstrate the regret upper bound of FORKS as follows

\begin{theorem}[Regret Bound of FORKS]
    \label{thm:OnlineKernel:Sketch:SJLT:regret}
        Let \(\delta_0, \epsilon_0 \in (0, 1)\). 
        Set the number of submatrices in SJLT as \(D = \Theta(\log^3(s_m))\).
        Suppose the parameters for updating \(\bm{A}_t\) in FORKS satisfy \(\eta_i = 0\) and \(\sigma_i \ge L_{\mathrm{Cur}} > 0\). 
        Assume the eigenvalues of \(\bm{K}\) decay polynomially with decay rate \(\beta > 1\), and that the SJLT \(\bm{S}_p\) satisfies Assumption~\ref{assum:FORKS:Matrix} with failure probabilities \(\delta_1, \delta_2 \in (0, 1)\). 
        If the sketch sizes of $\bm S_p$ and $\bm S_m$ satisfy
        \begin{align*}
           s_p = \mathrm{\Omega}\left( s_m~\mathrm{polylog} ( s_m \delta_0^{-1} ) / \epsilon_0^2 \right),~
           s_m = \Omega ( C_\mathrm{Coh} k \log k).
        \end{align*}
        Then, with probability at least $1 - \delta$,
        \begin{align*}
            \mathrm{Reg}_T (f^*)
            &\leq
            \frac{\alpha D_{\bm w}^2}{2}  +
            \frac{k}{2L_{\mathrm{Cur}}}  O\left(\log T\right) +
            \frac{\lambda}{2} \| f^* \|_{\mathcal{H}_{\kappa}}^2+\\
            &\hphantom{{}={}} 
            \dfrac{\sqrt{1 + \epsilon}}{\lambda}O\left(\sqrt{B}\right) + \dfrac{(k+1)^{-\beta}}{2\lambda\theta}+
            \\
            &\hphantom{{}={}} 
            \dfrac{1}{\lambda(\beta-1)}
            \left( \dfrac{3}{2} -  \dfrac{B+ \lfloor (T-B)/\rho \rfloor}{T}\right).
        \end{align*}
        where $\delta = \delta_0 + \delta_1 + \delta_2$, \(D_{\bm{w}}\) denotes the diameter of the weight vector space of the hypothesis on the incremental sketches, and \(\epsilon\) is defined as:
        \[
        \sqrt{\epsilon} = 2\gamma \sqrt{\frac{T}{\delta_1 \delta_2}} + \sqrt{\frac{2\gamma}{\delta_2}} \left( \epsilon_0^2 + 2\epsilon_0 + 2 \right),
        \]
        with \(\gamma = s_m / s_p\).
\end{theorem}

\begin{proof}[proof sketch]
Let $\bm w^*$ denote the optimal hypothesis on the incremental sketches in hindsight, we decompose the instantaneous regret $\ell_t (\bm w_t) - \ell_t (\bm w^*)$ into two terms as follows
\begin{align*}
        \underbrace{\ell_t (\bm w_t) - \ell_t (\bm w_t^*)}_{\text{Term 1: Optimization Error}} +
        \underbrace{\ell_t (\bm w_t^*) - \ell_t (\bm w^*)}_{\text{Term 2: Estimation Error}}.
\end{align*}
The optimization error stems from the optimization step of second-order online gradient descent, while the estimation error arises from the incremental matrix sketching and the truncated singular values in TISVD. 
We provide upper bounds for the errors and present the detailed proof in Appendix C.1.

\end{proof}

\begin{remark}[Assumption of polynomial decay]
\label{rem:regret}
The assumption of polynomial eigenvalue decay for the kernel matrix is widely applicable and holds for various types of kernels, including shift-invariant, finite-rank, and convolution kernels~\cite{liu2015eigenvalues,belkin2018approximation}. 
This decay property guarantees that the accumulated truncated singular values of TISVD can be upper bounded by the number of update rounds.
\end{remark}

\begin{remark}[Convex case]
\label{rem:convex}
Note that when $L_{\mathrm{Cur}} = 0$, Assumption~\ref{assm:OKL:Curvature} essentially enforces convexity.
In the worst case when $L_{\mathrm{Cur}} = 0$, the regret bound in the convex case degenerates to $O(\sqrt{T})$. 
The detailed proof is included in Appendix C.2.
\end{remark}

By configuring the update cycle as \(\rho = \lfloor \theta (T-B) \rfloor\), where \(\theta \in (0, 1)\), and setting the sketch size ratio to \(\gamma = O(\log(T) / \sqrt{T})\), we derive an upper bound on regret of \(O(T^{\frac{1}{4}} \log T)\) for second-order online kernel learning. In contrast, the regret bound for first-order online kernel learning is \(O(\sqrt{T})\), while requiring a budget of \(B = O(T)\)~\cite{lu2016large}, making it less favorable than FORKS.

Under the assumption that the eigenvalues of the kernel matrix decay polynomially with rate \(\beta\), the regret bound for the existing second-order online kernel learning algorithm PROS-N-KONS~\cite{calandriello2017efficient} is \(O(T^{\frac{1}{\beta}} \log^2 T)\), which is highly sensitive to the eigenvalue decay rate $\beta$.
In practical streaming scenarios, however, ensuring a sufficiently rapid decay of the kernel matrix is often unrealistic. 
When the properties of the streaming matrix are less favorable, such as in the case where \(\beta < 4\), the regret of PROS-N-KONS can be worse than that of FORKS.

\section{Experiments}
\label{sec: Experiments}
In this section, we conduct experiments to evaluate the performance of FORKS on several datasets. 
The details of datasets and experimental setup are presented in Appendix D.1, D.2.

\begin{table*}[htp]
\footnotesize
\begin{spacing}{.95}
\centering
\resizebox{0.75\linewidth}{!}{
\begin{tabular}{lcccccc}
\toprule
     \multirow{2}{*}{Algorithm}
             & \multicolumn{2}{c}{\texttt{german}}   & \multicolumn{2}{c}{\texttt{svmguide3}}
             & \multicolumn{2}{c}{\texttt{spambase}} \\\cmidrule(lr){2-3} \cmidrule(lr){4-5} \cmidrule(lr){6-7}
 & Mistake rate  &  Time  & Mistake rate  &  Time  & Mistake rate  &  Time  \\\midrule
RBP            &38.830 	$\pm$ 0.152 	&  0.003 	 &29.698 	$\pm$ 1.644 	&  0.003     &35.461 	$\pm$ 0.842 	&  0.025 \\
BPA-S          &35.235 	$\pm$ 0.944 	&  0.004 	 &29.027 	$\pm$ 0.732 	&  0.004     &34.394 	$\pm$ 2.545 	&  0.039 \\
Projectron     &36.875 	$\pm$ 1.403 	&  0.003 	 &25.060 	$\pm$ 0.373 	&  0.003     &32.659 	$\pm$ 0.914 	&  0.031 \\
BOGD           &33.705 	$\pm$ 1.446 	&  0.007 	 &29.904 	$\pm$ 1.653 	&  0.006     &32.859 	$\pm$ 0.478 	&  0.049 \\
FOGD           &30.915 	$\pm$ 0.845 	&  0.025 	 &30.024 	$\pm$ 0.787 	&  0.022     &\textbf{25.651} 	$\pm$ \textbf{0.349} 	&  0.175 \\
NOGD           &26.715 	$\pm$ 0.552 	&  0.014 	 &\underline{19.964 	$\pm$ 0.077} 	&  0.008     &31.003 	$\pm$ 0.751 	&  0.077 \\
SkeGD          &\textbf{25.170} 	$\pm$ \textbf{0.391} 	&  0.009 	 &19.976 	$\pm$ 0.105 	&  0.007     &32.413 	$\pm$    1.886 	&  0.067 \\
\midrule
PROS-N-KONS    &31.235 	$\pm$ 0.939 	&  1.017 	 &24.529 	$\pm$ 0.561 	&  0.015     &32.227 	$\pm$ 0.678 	&  6.638 \\
FORKS (Ours)       &\underline{26.425 	$\pm$ 0.562} 	&  0.008 	 &\textbf{19.710} 	$\pm$ \textbf{0.557} 	&  0.009     &\underline{30.662 	$\pm$ 0.670} 	&  0.070 \\
\midrule
     \multirow{2}{*}{Algorithm}
             & \multicolumn{2}{c}{\texttt{codrna}}   & \multicolumn{2}{c}{\texttt{w7a}}
             & \multicolumn{2}{c}{\texttt{ijcnn1}} \\\cmidrule(lr){2-3}\cmidrule(lr){4-5}\cmidrule(lr){6-7}
 & Mistake rate  &  Time  & Mistake rate  &  Time  & Mistake rate  &  Time  \\\midrule
RBP            &22.644 	$\pm$ 0.262 	&  0.210 	 &5.963 	$\pm$ 0.722 	&  0.945     &21.024     $\pm$ 0.578 	&  0.633 \\
BPA-S          &17.029 	$\pm$ 0.303 	&  0.313 	 &3.001     $\pm$ 0.045 	&  1.145     &11.114     $\pm$ 0.064 	&  0.747 \\
Projectron     &19.257 	$\pm$ 4.688 	&  0.341 	 &3.174 	$\pm$ 0.014 	&  0.965     &\ \ 9.478 	$\pm$ 0.001 	&  0.621 \\
BOGD           &17.305 	$\pm$ 0.146 	&  0.507 	 &3.548 	$\pm$ 0.164 	&  0.970     &11.559 	$\pm$ 0.174 	&  0.724 \\
FOGD           &\underline{13.103 	$\pm$ 0.105} 	&  1.480 	 &2.893 	$\pm$ 0.053 	&  2.548     &9.674 	$\pm$ 0.105 	&  3.125 \\
NOGD           &17.915 	$\pm$ 3.315 	&  0.869 	 &\underline{2.579 	$\pm$ 0.007} 	&  2.004     &\ \ \textbf{9.379} 	$\pm$ \textbf{0.001} 	&  1.457 \\
SkeGD          &13.274 	$\pm$ 0.262 	&  0.779 	 &2.706 	$\pm$ 0.335 	&  2.093     &11.898 	$\pm$ 1.440 	&  2.216 \\
\midrule
PROS-N-KONS    &13.387 	$\pm$ 0.289 	&  114.983 	 &3.016 	$\pm$ 0.007 	&  92.377     &\ \ 9.455 	$\pm$ 0.001 	&  5.000 \\
FORKS (Ours)       &\textbf{12.795} 	$\pm$ \textbf{0.360} 	&  0.918 	 &\textbf{2.561}     $\pm$ \textbf{0.038} 	&  2.240     &\ \ \underline{9.381 $\pm$ 0.001} 	&  2.480 \\
\bottomrule
\end{tabular}
}
\caption{Comparisons among first-order algorithms RBP, BPA-S, BOGD, Projectron, NOGD, SkeGD, FOGD and second-order algorithms PROS-N-KONS, FORKS w.r.t. the mistake rates (\%) and the running time (s). The best result is highlighted in $\textbf{bold}$ font, and the second best result is \underline{underlined}.}
\vspace{-1em}
\label{tbl: OL_new}
\end{spacing}
\end{table*}


\subsection{Experiments Under a Fixed Budget}
In this section, we demonstrate the performance of FORKS under a fixed budget, employing six widely recognized classification benchmarks.
We compare FORKS with the existing budgeted-based online learning algorithms, including first-order algorithms RBP~\cite{cavallanti2007tracking}, BPA-S~\cite{wang2010online}, BOGD~\cite{zhao2012fast}, FOGD, NOGD~\cite{lu2016large}, Projectron~\cite{orabona2008projectron}, SkeGD~\cite{zhang2019incremental} and second-order algorithm PROS-N-KONS~\cite{calandriello2017efficient}.
All algorithms are trained using hinge loss, and their performance is measured by the average online mistake rate.

For all the algorithms, we set a fixed budget $B=50$ for small datasets $(N\le10000)$ and $B=100$ for large datasets.
Furthermore, we set buffer size $\Tilde{B} = 2B, \gamma = 0.2$, $s_p = B$, $s_m = \gamma{s_p}$, $\theta = 0.3$, and update cycle $\rho=\lfloor\theta{N}\rfloor$ in SkeGD and FORKS if not specially specified.
For algorithms with rank-$k$ approximation, we uniformly set $k=0.1B$.
Besides, we use the same experimental settings for FOGD (feature dimension = 4B).
The results are presented in Table~\ref{tbl: OL_new}.
Our FORKS shows the best performance on \texttt{svmguide3}, \texttt{codrna}, \texttt{w7a}, and the suboptimal performance on other datasets.
The update time of FORKS is comparable to that of the majority of first-order algorithms, including NOGD and SkeGD.
Besides, FORKS is significantly more efficient than the existing second-order method PROS-N-KONS in large-scale datasets such as \texttt{codrna} and \texttt{w7a}.

\begin{figure}[htp]
	\centering
	\begin{subfigure}{0.49\linewidth}
		\centering
		\includegraphics[width=\linewidth]{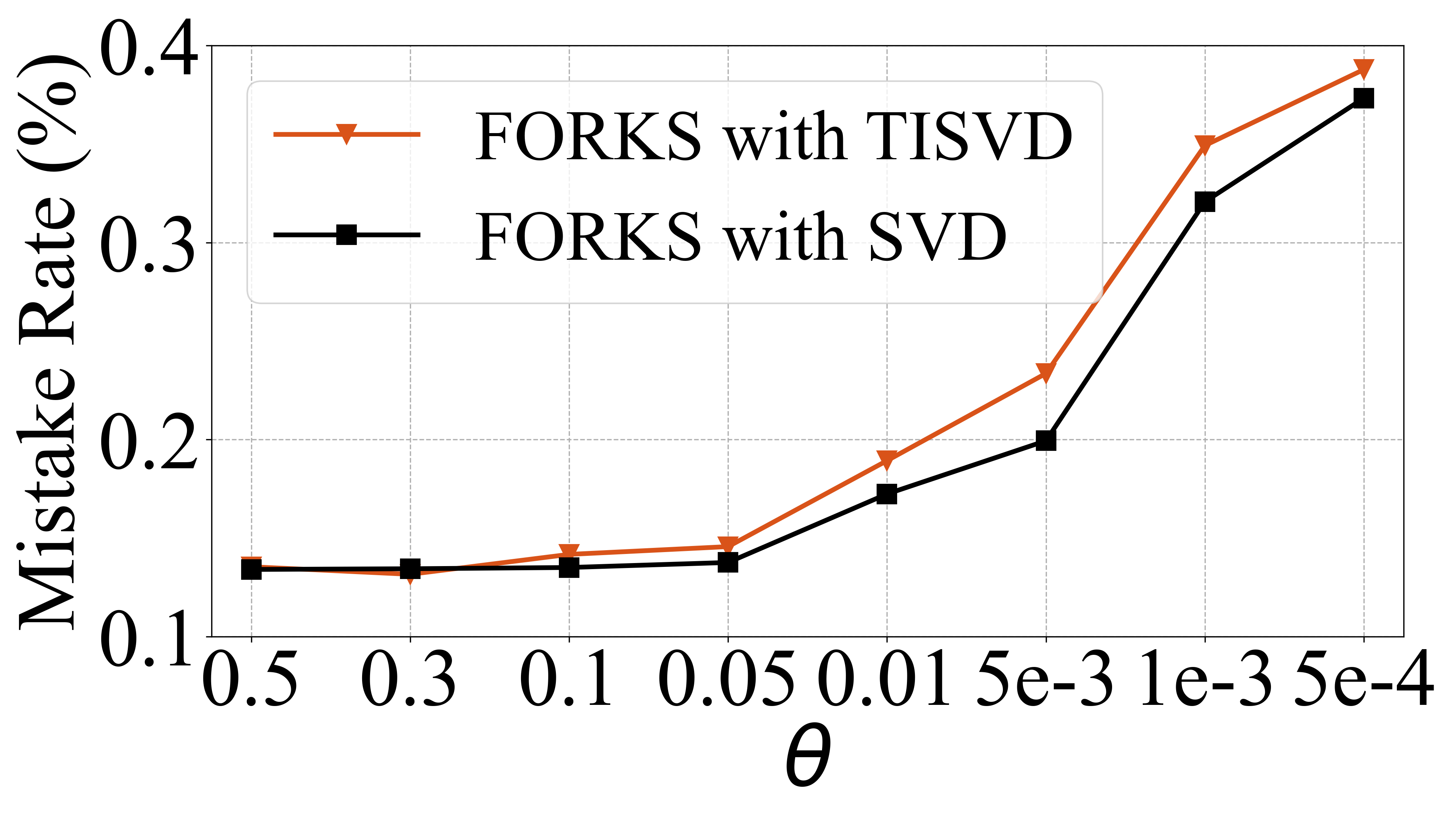}
		\caption{average mistake rate}
		\label{TISVD_exp_1}
	\end{subfigure}
	\centering
	\begin{subfigure}{0.49\linewidth}
		\centering
		\includegraphics[width=\linewidth]{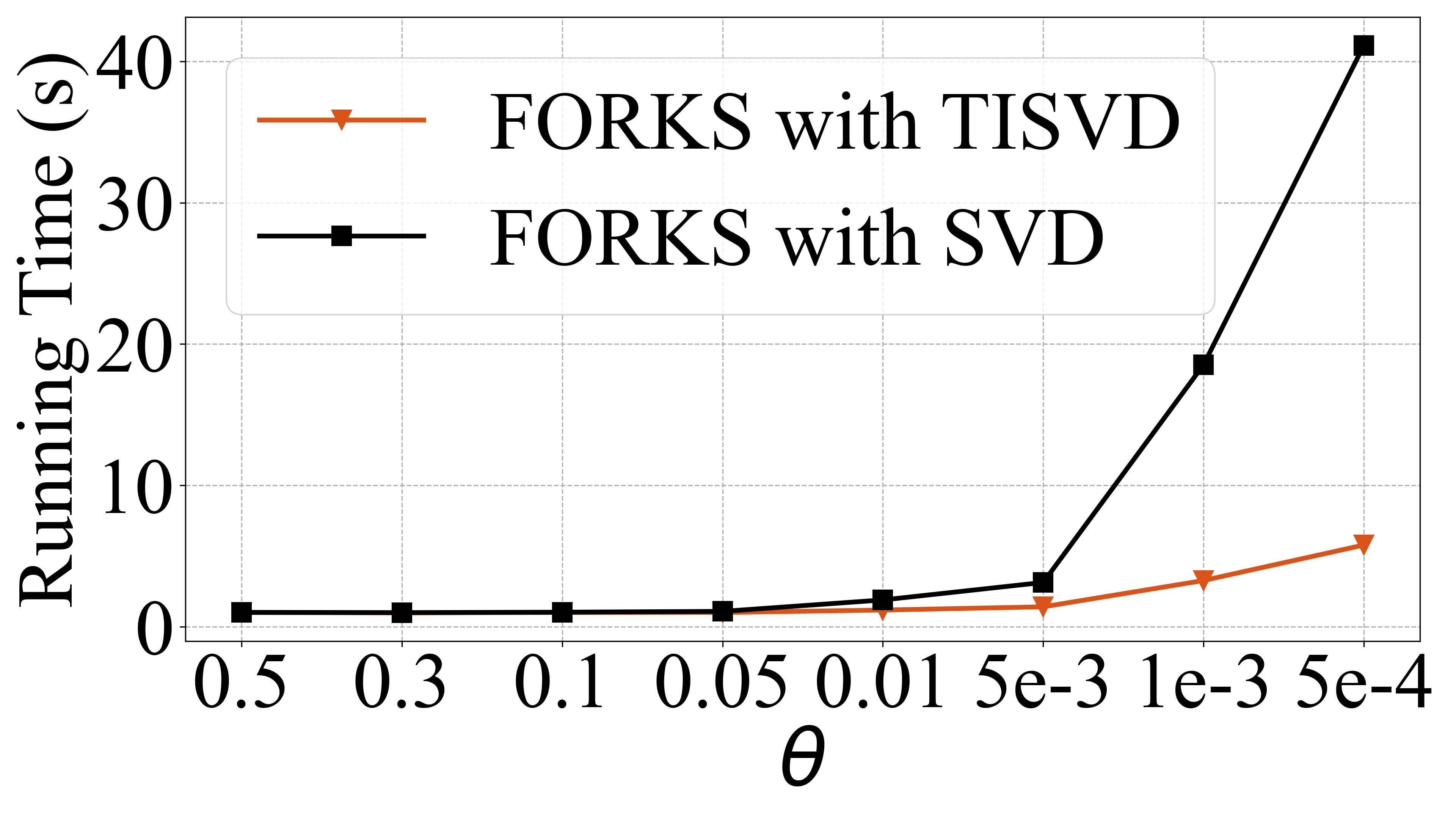}
		\caption{average running time}
		\label{TISVD_exp_2}
	\end{subfigure}
	\caption{The average mistake rates and average running time w.r.t. TISVD on \texttt{codrna}.
    \vspace{-1em}}
	\label{TISVD_exp}
\end{figure}

Then, we conduct experiments to evaluate how TISVD affects the performance of the algorithm.
We use the same experimental setup in \texttt{codrna} and vary the update rate $\theta$ from 0.5 to 0.0005.
Figure~\ref{TISVD_exp} demonstrates that TISVD maintains efficient decomposition speed without excessively reducing performance.
Considering that frequent updates can potentially result in an elevated loss, it is essential to carefully choose an optimal update cycle that strikes a balance between achieving superior accuracy and maintaining efficiency.

\subsection{Experiments Under Adversarial Environment}
\label{sec: Experiments under Adversarial Environment}
To empirically validate the algorithms under an adversarial environment, we build adversarial datasets using the benchmark \texttt{codrna} and \texttt{german}.
We compare FORKS with first-order algorithms BOGD~\cite{zhao2012fast}, SkeGD~\cite{zhang2019incremental}, NOGD~\cite{lu2016large} and second-order algorithm PROS-N-KONS~\cite{calandriello2017efficient} under the same budget $B=200$.
Besides, we set $\gamma = 0.2$, $s_p = 0.75B$, $s_m = \gamma{s_p}, k=0.1B$ and update cycle $\rho=\lfloor0.005(N-B)\rfloor$ in SkeGD and FORKS.
Inspired by the adversarial settings in~\cite{calandriello2017efficient,wang2018minimizing,zhang2019incremental}, we generate an online learning game with $b$ blocks.
At each block, we extract an instance from the dataset and repeat it for $r$ rounds.
In addition, the labels are flipped in each even block by multiplying them with -1.
We set $b=500$, $r=10$ for \texttt{codrna-1} and \texttt{german-1}.

\begin{table}[h]
\footnotesize
\centering
\begin{spacing}{1}
\resizebox{1.0\linewidth}{!}{
\begin{tabular}{lcccc}
\toprule
     \multirow{2}{*}{Algorithm}
             & \multicolumn{2}{c}{\texttt{codrna-1}}   & \multicolumn{2}{c}{\texttt{german-1}}
             \\\cmidrule(lr){2-3} \cmidrule(lr){4-5}
 & Mistake rate  &  Time  & Mistake rate  &  Time  \\\midrule
BOGD           &26.066 	$\pm$ 1.435 	&  0.029 	 &32.131 	$\pm$ 1.079 	&  0.042    \\
NOGD           &29.780 	$\pm$ 1.257 	&  0.024 	 &28.103 	$\pm$ 1.247 	&  0.040  \\
SkeGD          &24.649 	$\pm$ 5.087 	&  0.269 	 &11.026 	$\pm$ 4.018 	&  0.113  \\
PROS-N-KONS    &21.299 	$\pm$ 1.364 	&  3.323 	 &17.174 	$\pm$ 1.437 	&  0.477     \\
FORKS (Ours)       &\ \ \textbf{6.752} 	$\pm$ \textbf{1.647} 	&  0.023 	 &\ \ \textbf{5.142} 	$\pm$ \textbf{0.215} 	&  0.035   \\
\bottomrule
\end{tabular}
}
\caption{Comparisons among BOGD, NOGD, PROS-N-KONS, SkeGD and our FORKS w.r.t. the mistake rates (\%) and the running time (s). The best result is highlighted in $\textbf{bold}$ font.}
\vspace{-1em}
\label{tbl: Ad}
\end{spacing}
\end{table}

Experimental results are presented in Table~\ref{tbl: Ad}. 
It is observed that in the adversarial environment, the performance of all methods significantly decreases with the increase of adversarial changes except for FORKS.
This is due to the fact that FORKS accurately captures the concept drifting through the incremental update of the sketch matrix and the execution of rapid second-order gradient descent.
Moreover, FORKS maintains its efficiency comparable to first-order algorithms, thereby ensuring that improved performance is achieved without sacrificing computational time.

\subsection{Experiments on Large-Scale Datasets}
\label{sec: Experiments on Large-Scale Datasets}
In this experiment, we evaluate the efficiency and effectiveness of FORKS on streaming recommendation.
We use \texttt{KuaiRec}, which is a real-world dataset collected from the recommendation logs of the video-sharing mobile app Kuaishou~\cite{gao2022kuairec}.
We conduct experiments on the dense matrix of \texttt{KuaiRec}, which consists of $4,494,578$ instances with associated timestamps, making it an ideal benchmark for evaluating large-scale online learning tasks.
We test the performance of the algorithm used in Section 5.3 under different budgets $B$ ranging from $100$ to $500$.
To avoid excessive training time, we use a budgeted version of PROS-N-KONS that stops updating the dictionary at a maximum budget of $B_{\max}=100$. 
Since the buffer size of PROS-N-KONS is data-dependent, we repeat the training process 20 times to compute the average error rate and the average time for comparison.
In addition to the hinge loss, we use squared hinge loss to evaluate the performance.

\begin{figure}[htp]
	\centering
	\begin{subfigure}{0.49\linewidth}
		\centering
		\includegraphics[width=\linewidth]{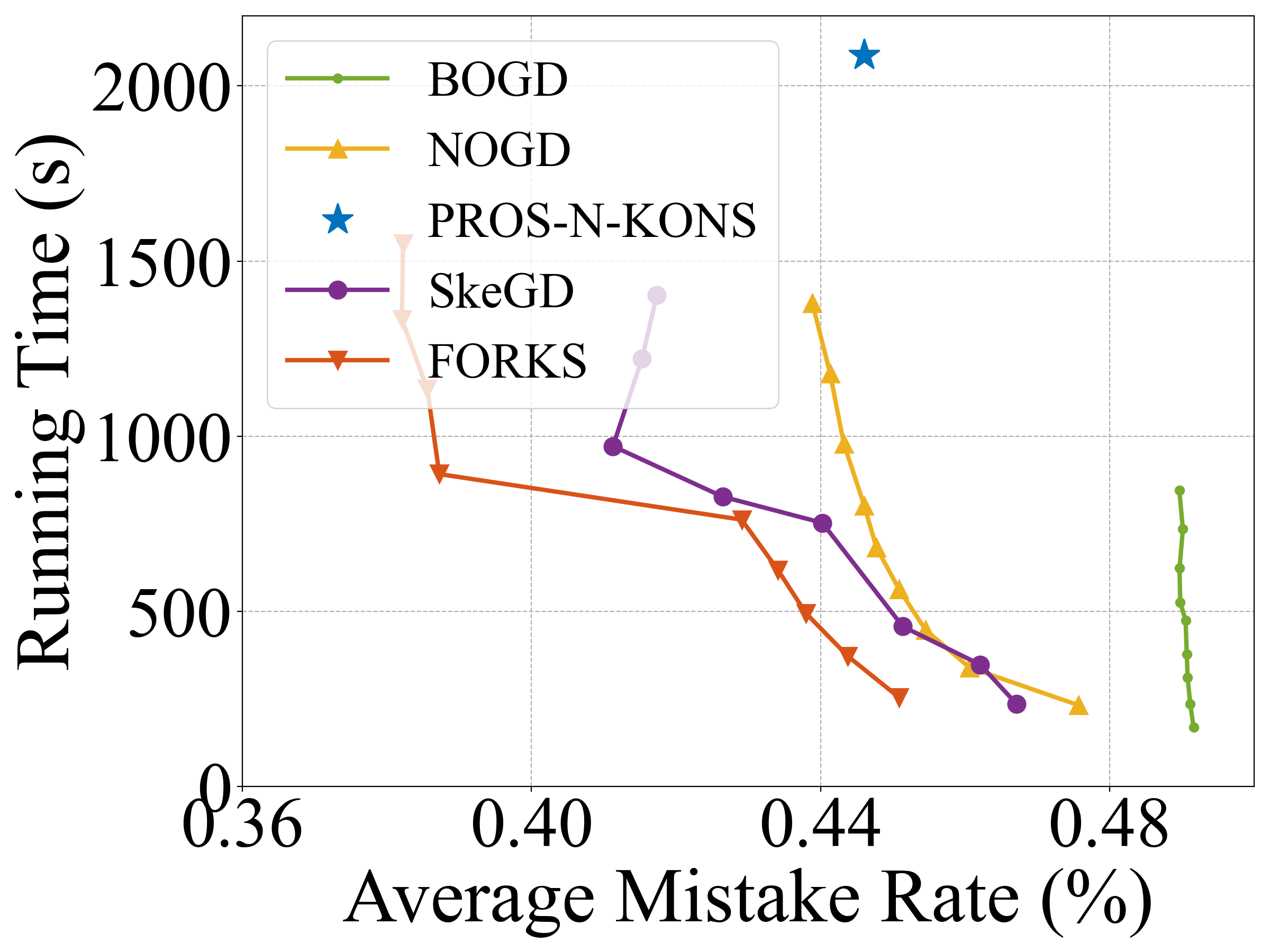}
		\caption{Hinge loss}
		\label{Kuai_error}
	\end{subfigure}
	\centering
	\begin{subfigure}{0.49\linewidth}
		\centering
		\includegraphics[width=\linewidth]{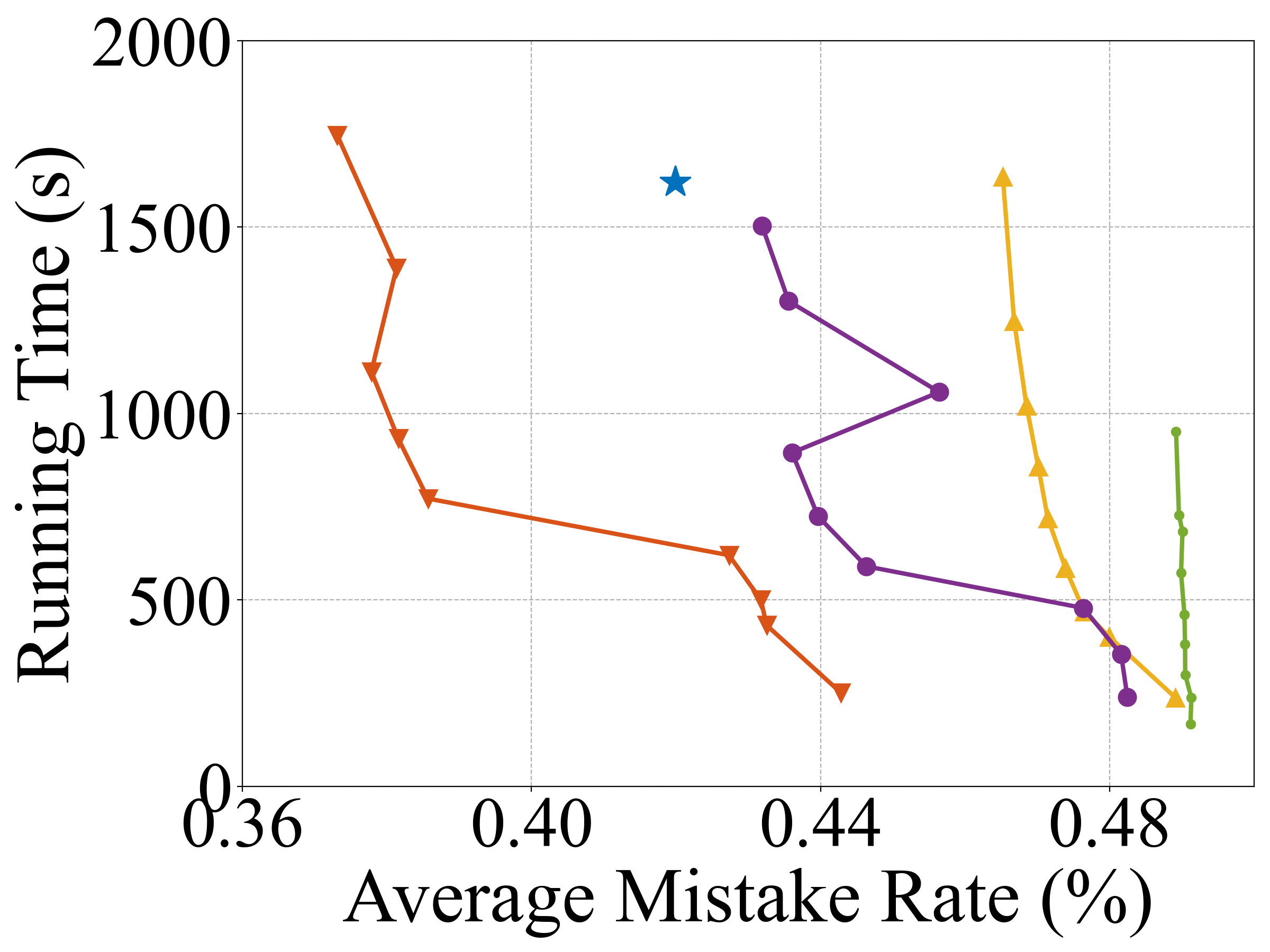}
		\caption{Squared hinge loss}
		\label{Kuai_time}
	\end{subfigure}
	\caption{The tradeoff between running time and the average mistake rate on \texttt{KuaiRec}. As PROS-N-KONS utilizes an adaptive budget, thereby being depicted as a single point in the figures.
    \vspace{-1em}}
    \label{Kuai_exp}
\end{figure}

Figure~\ref{Kuai_exp}~(a) shows the tradeoff between running time and the average mistake rate in the experiment using hinge loss. 
Figure~\ref{Kuai_exp}~(b) shows the tradeoff between running time and the average mistake rate in the experiment using squared hinge loss. 
We observe that FORKS consistently achieves superior learning performance while maintaining comparable time costs to the other first-order algorithms, regardless of the loss function's shape.
In particular, for squared hinge loss, both PROS-N-KONS and FORKS significantly outperform first-order models, highlighting the advantages of second-order methods under exp-concave losses.
We also observe that FORKS demonstrates considerably higher efficiency than the second-order algorithm PROS-N-KONS. Under squared hinge loss, to achieve a comparable online error rate, FORKS requires only approximately 500 seconds, while PROS-N-KONS takes over 1500 seconds, resulting in a threefold speedup.

\section{Conclusion}
This paper introduces FORKS, a fast second-order online kernel learning approach. By leveraging incremental matrix sketching and decomposition techniques, FORKS efficiently addresses the computational challenges inherent in kernel feature mapping and hypothesis updates. The proposed method achieves a logarithmic regret bound while maintaining linear time complexity relative to the budget, significantly improving the efficiency of existing second-order methods. Extensive experimental evaluations on various datasets demonstrate the superior scalability and robustness of our approach.


\newpage

\section*{Acknowledgments}
This research was supported in part by National Science and Technology Major Project (2022ZD0114802), by National Natural Science Foundation of China (No. 92470128, No. U2241212, No. 62376275), by Beijing Outstanding Young Scientist Program No.BJJWZYJH012019100020098, by Huawei-Renmin University joint program on Information Retrieval. We also wish to acknowledge the support provided by the fund for building world-class universities (disciplines) of Renmin University of China, by Engineering Research Center of Next-Generation Intelligent Search and Recommendation, Ministry of Education, by Intelligent Social Governance Interdisciplinary Platform, Major Innovation $\&$ Planning Interdisciplinary Platform for the “Double-First Class” Initiative, Public Policy and Decision-making Research Lab, and Public Computing Cloud, Renmin University of China.
The work was partially done at Beijing Key Laboratory of Research on Large Models and Intelligent Governance, MOE Key Lab of Data Engineering and Knowledge Engineering, Engineering Research Center of Next-Generation Intelligent Search and Recommendation, MOE, and Pazhou Laboratory (Huangpu), Guangzhou, Guangdong 510555, China.

\appendix

\newpage
\section{Related Work}

In this section, we review the related work on budget-based and sampling-based online kernel learning, as well as randomized matrix sketching algorithms.

\paragraph{Budget-based Online Kernel Learning.}

Kernelizing the online gradient descent may lead to linear growth in computation time with the number of rounds, thus significantly increasing the hypothesis update time. Budget-based methods mitigate this by maintaining a fixed-size buffer of support vectors, which limits the model size and reduces computational complexity. 
Various budget maintenance strategies have been proposed to control the buffer size. \cite{DekelSS05} proposed discarding the oldest support vector to maintain the budget, while \cite{orabona2008projectron} reduced buffer size by projecting new examples onto the linear span of the existing support vectors in the feature space.
However, the buffer of SVs cannot be used for kernel matrix approximation directly.

\paragraph{Sampling-based Online Kernel Learning.}

In contrast to budget-based methods, sampling techniques have been introduced into online kernel learning to incrementally approximate the kernel. 
NOGD utilizes the Nystr\"{o}m method for kernel matrix approximation and constructs an explicit feature mapping through singular value decomposition~\cite{lu2016large}. SKETCHED-KONS is a second-order online kernel learning approach that employs online sampling techniques. However, it applies budgeting strategies only to the Hessian of the second-order updates, not to the hypothesis itself, leading to a time-consuming evaluation process~\cite{calandriello2017second}. 
PROS-N-KONS further enhances second-order updates for hypothesis updating, achieving a more efficient per-step cost that scales with the effective dimension~\cite{calandriello2017efficient}.

\paragraph{Randomized Matrix Sketching.}
It is a matrix transformation technique performed in an online manner, without the need to store the entire streaming matrix. The sketch matrix can effectively approximate the streaming matrix due to the property of matrix product preservation. Several sketching strategies have been proposed, including Count Sketch~\cite{charikar2002finding}, the Sparser Johnson-Lindenstrauss Transform~\cite{kane2014sparser}, and sampling~\cite{Woodruff2014SAA}. 
In offline learning, randomized matrix sketching is employed to reduce the complexity of solving the kernel optimization problem~\cite{wang2016spsd}.SkeGD further incorporates randomized matrix sketching into online kernel learning, formulating a first-order method~\cite{zhang2019incremental}.
However, extending it to the second-order gradient updates is non-trivial as it necessitates a new analytical framework to investigate the sketching properties under the directional curvature.

\section{Omitted Details for Section~\ref{sec: FORKS: The Proposed Algorithm}}

In this section, we provide the omitted details for Section~\ref{sec: FORKS: The Proposed Algorithm}. 
We begin by introducing the incremental maintenance of the randomized sketch in Appendix~\ref{Incremental Maintenance of Randomized Sketch}, which is a key operation for approximating the kernel matrix in an online setting. Next, we present the pseudo-code for the proposed TISVD in Appendix~\ref{TISVD code} and discuss the distinctions between TISVD and the efficient RSVD method in Appendix~\ref{TISVD Discussion}.

\subsection{Incremental Maintenance of Randomized Sketch}
\label{Incremental Maintenance of Randomized Sketch}
At round $t + 1$, a new example $\bm{x}_{t+1}$ arrives, and the kernel matrix $\bm{K}^{(t+1)}\in\mathbb{R}^{(t+1)\times(t+1)}$ can be represented as a bordered matrix and approximated using several small sketches as follows:
\begin{align*}
    \bm{K}^{(t+1)} &= \begin{bmatrix}
    \bm{K}^{(t)}&\boldsymbol{\psi}^{(t+1)}\\
    \boldsymbol{\psi}^{(t+1)\top}& \kappa{(\bm{x}_{t+1},\bm{x}_{t+1})}
\end{bmatrix} \\
&\approx
\bm{C}_m^{(t+1)}
\left(\boldsymbol{\Phi}_{pm}^{(t+1)}\right)^{\dagger}\boldsymbol{\Phi}_{pp}^{(t+1)}\left(\boldsymbol{\Phi}_{pm}^{(t+1)\top}\right)^{\dagger}
\bm{C}_m^{(t+1)\top},
\end{align*}
where $\boldsymbol{\psi}^{(t+1)} = [\kappa{(\bm{x}_{t+1},\bm{x}_1)}, \kappa{(\bm{x}_{t+1},\bm{x}_2)},\ldots,\kappa{(\bm{x}_{t+1},\bm{x}_t)}]^{\top}$. 
The sketches can be represented as
\begin{equation*}
    \begin{aligned}
    \boldsymbol{\Phi}_{pm}^{(t+1)} = \bm{S}_p^{(t+1)\top}\bm{C}_m^{(t+1)},~
    \boldsymbol{\Phi}_{pp}^{(t+1)} = \bm{S}_p^{(t+1)\top}\bm{C}_p^{(t+1)},
\end{aligned}
\end{equation*}
where $\bm{C}_m^{(t+1)} = \bm{K}^{(t+1)}\bm{S}_m^{(t+1)}$, $\bm{C}_p^{(t+1)} = \bm{K}^{(t+1)}\bm{S}_p^{(t+1)}$.

The sketches are obtained using an SJLT 
$\bm{S}_p^{(t+1)} \in \mathbb{R}^{(t+1)\times{s_p}}$ and a column-sampling matrix $\bm{S}_m^{(t+1)} \in \mathbb{R}^{(t+1)\times{s_m}}$.
We partition the sketch matrices into block matrices as $\bm{S}_p^{(t+1)}=\left[\bm{S}_p^{(t)\top}, \bm{s}_p^{(t+1)}\right]^{\top}$, $\bm{S}_m^{(t+1)}=\left[\bm{S}_m^{(t)\top}, \bm{s}_m^{(t+1)}\right]^{\top}$, where $\bm{s}_m^{(t+1)}\in{\mathbb{R}^{s_m}}$ is a sub-sampling vector and $\bm{s}_p^{(t+1)}\in{\mathbb{R}^{s_p}}$ is a new row vector of $\bm{S}_p^{(t+1)}$ sharing the same hash functions.

Furthermore,  we can update the sketches $\boldsymbol{\Phi}_{pm}^{(t+1)}$ and $\boldsymbol{\Phi}_{pp}^{(t+1)}$ using rank-1 modifications as follows:
The sketch  $\bm \Phi^{(t+1)}_{pm} $ can be maintained as
    \begin{align*}
    &\hphantom{{}={}}    
        \bm \Phi^{(t+1)}_{pm}\\
        &=
        {\bm S}_{\mathrm{p}}^{(t+1)\top} {\bm C}_{\mathrm{m}}^{(t+1)}\\
        &=
        {\bm S}_{\mathrm{p}}^{(t+1)\top} {\bm K}^{(t+1)} {\bm S}_{\mathrm{m}}^{(t+1)}
        \\
        &=
        [{\bm S}_{\mathrm{p}}^{(t)\top}, {\bm s}_{\mathrm{p}}^{(t+1)}]
          \left[
            \begin{array}{c c}
              {\bm K}^{(t)}               & {\bm \psi }^{(t+1)} \\
              {\bm \psi }^{(t+1)\top}  & \kappa{(\bm{x}_{t+1},\bm{x}_{t+1})} \\
            \end{array}
          \right]
          \left[
            \begin{array}{c}
              {\bm S}_{\mathrm{m}}^{(t)} \\
              {\bm s}_{\mathrm{m}}^{(t+1)\top} \\
            \end{array}
          \right]
        \\
        &=
        \left[
            \begin{array}{c}
            {\bm S}_{\mathrm{p}}^{(t)\top} {\bm K}^{(t)} + {\bm s}_{\mathrm{p}}^{(t+1)} {\bm \psi }^{(t+1)\top}\\
            {\bm S}_{\mathrm{p}}^{(t)\top} {\bm \psi }^{(t+1)} + \kappa{(\bm{x}_{t+1},\bm{x}_{t+1})} {\bm s}_{\mathrm{p}}^{(t+1)}\\
            \end{array}
        \right]^{\top}
        \left[
            \begin{array}{c}
              {\bm S}_{\mathrm{m}}^{(t)} \\
              {\bm s}_{\mathrm{m}}^{(t+1)\top} \\
            \end{array}
        \right]\\
        &=
        {\bm S}_{\mathrm{p}}^{(t)\top} {\bm K}^{(t)} {\bm S}_{\mathrm{m}}^{(t)} +
         {\bm R}_{pm}^{(t+1)} + {\bm R}_{\mathrm{mp}}^{(t+1)\top}
        +  {\bm T}_{pm}^{(t+1)}\\
        &=
        \bm \Phi^{(t)}_{pm}  +
         {\bm R}_{pm}^{(t+1)} + {\bm R}_{\mathrm{mp}}^{(t+1)\top}
        +  {\bm T}_{pm}^{(t+1)},
    \end{align*}
    where the modifications are performed using the following three rank-1 matrices
    \begin{align*}
        {\bm R}_{pm}^{(t+1)} &= {\bm s}_{\mathrm{p}}^{(t+1)} {\bm \psi }^{(t+1)\top} {\bm S}_{\mathrm{m}}^{(t)}, \\
        {\bm R}_{\mathrm{mp}}^{(t+1)} &= {\bm s}_{\mathrm{m}}^{(t+1)} {\bm \psi }^{(t+1)\top} {\bm S}_{\mathrm{p}}^{(t)}, \\
        {\bm T}_{pm}^{(t+1)} &= \kappa{(\bm{x}_{t+1},\bm{x}_{t+1})} {\bm s}_{\mathrm{p}}^{(t+1)} {\bm s}_{\mathrm{m}}^{(t+1)\top}.
    \end{align*}
    For sketch $\bm \Phi^{(t+1)}_{pp}$, we have
    \begin{align*}
       &\hphantom{{}={}}
       \bm \Phi^{(t+1)}_{pp}\\
       &=
       {\bm S}_{\mathrm{p}}^{(t+1)\top} {\bm C}_{\mathrm{p}}^{(t+1)}\\
       &=
       {\bm S}_{\mathrm{p}}^{(t+1)\top} {\bm K}^{(t+1)} {\bm S}_{\mathrm{p}}^{(t+1)}
       \\
       &=
        [{\bm S}_{\mathrm{p}}^{(t)\top}, {\bm s}_{\mathrm{p}}^{(t+1)}]
          \left[
            \begin{array}{c c}
              {\bm K}^{(t)}               & {\bm \psi }^{(t+1)} \\
              {\bm \psi }^{(t+1)\top}  & \kappa{(\bm{x}_{t+1},\bm{x}_{t+1})} \\
            \end{array}
          \right]
          \left[
            \begin{array}{c}
              {\bm S}_{\mathrm{p}}^{(t)} \\
              {\bm s}_{\mathrm{p}}^{(t+1)\top} \\
            \end{array}
          \right]
        \\
        &=
        \left[
            \begin{array}{c}
                {\bm S}_{\mathrm{p}}^{(t)\top} {\bm K}^{(t)} + {\bm s}_{\mathrm{p}}^{(t+1)} {\bm \psi }^{(t+1)\top}\\
                {\bm S}_{\mathrm{p}}^{(t)\top} {\bm \psi }^{(t+1)} + \kappa{(\bm{x}_{t+1},\bm{x}_{t+1})} {\bm s}_{\mathrm{p}}^{(t+1)}\\
            \end{array}
        \right]^{\top}
        \left[
            \begin{array}{c}
              {\bm S}_{\mathrm{p}}^{(t)} \\
              {\bm s}_{\mathrm{p}}^{(t+1)\top} \\
            \end{array}
        \right]
        \\
        &=
        {\bm S}_{\mathrm{p}}^{(t)\top} {\bm K}^{(t)} {\bm S}_{\mathrm{p}}^{(t)} +
         {\bm R}_{pp}^{(t+1)} + {\bm R}_{pp}^{(t+1)\top}
        +  {\bm T}_{pp}^{(t+1)},\\
        &=
       \bm \Phi^{(t)}_{pp} +
         {\bm R}_{pp}^{(t+1)} + {\bm R}_{pp}^{(t+1)\top}
        +  {\bm T}_{pp}^{(t+1)},
    \end{align*}
    where the modifications are done by the following two rank-$1$ matrices
    \begin{align*}
        {\bm R}_{pp}^{(t+1)} &= {\bm s}_{\mathrm{p}}^{(t+1)} {\bm \psi }^{(t+1)\top} {\bm S}_{\mathrm{p}}^{(t)}, \\
        {\bm T}_{pp}^{(t+1)} &= \kappa{(\bm{x}_{t+1},\bm{x}_{t+1})} {\bm s}_{\mathrm{p}}^{(t+1)} {\bm s}_{\mathrm{p}}^{(t+1)\top}.
    \end{align*}

In summary, sketches can be updated through low-rank matrices:
\begin{equation}
\label{UpdateSketch}
    \begin{aligned}
    \boldsymbol{\Phi}_{pm}^{(t+1)} &= \boldsymbol{\Phi}_{pm}^{(t)} + \bm{s}_p^{(t+1)}\boldsymbol{\psi}^{(t+1)\top}\bm{S}_m^{(t)} + \bm{S}_p^{(t)\top}\boldsymbol{\psi}^{(t+1)}\bm{s}_m^{(t+1)\top} \\ &\hphantom{{}={}} +\kappa{(\bm{x}_{t+1},\bm{x}_{t+1})}\bm{s}_p^{(t+1)}\bm{s}_m^{(t+1)\top} ,\\
    \boldsymbol{\Phi}_{pp}^{(t+1)} &= \boldsymbol{\Phi}_{pp}^{(t)} + 
    \bm{s}_p^{(t+1)}\boldsymbol{\psi}^{(t+1)\top}\bm{S}_p^{(t)}+
    \bm{S}_p^{(t)\top}\boldsymbol{\psi}^{(t+1)}\bm{s}_p^{(t+1)\top}\\
    &\hphantom{{}={}} +\kappa{(\bm{x}_{t+1},\bm{x}_{t+1})}\bm{s}_p^{(t+1)}\bm{s}_p^{(t+1)\top}.
    \end{aligned}
\end{equation}

Specifically, the proposed TISVD method efficiently constructs the time-varying explicit feature mapping $\boldsymbol{\phi}_t(\cdot)$ in \eqref{Qsvd} by setting $\bm{M} = \boldsymbol{\Phi}_{pp}$ and 
\begin{equation}
    \label{TISVD_for_FORKS}
    \begin{aligned}
        \bm{\Delta}_1 &= 
        \begin{bmatrix}
            \bm{s}_p^{(t+1)},   \bm{S}_p^{(t)\top}\boldsymbol{\psi}^{(t+1)},  \bm{s}_p^{(t+1)}
        \end{bmatrix}, \\
        \bm{\Delta}_2 &= 
        \begin{bmatrix}
             \bm{S}_p^{(t)\top}\boldsymbol{\psi}^{(t+1)}, \bm{s}_p^{(t+1)}, \kappa{(\bm{x}_{t+1},\bm{x}_{t+1})} \bm{s}_p^{(t+1)}
        \end{bmatrix}.
    \end{aligned}
\end{equation}

\subsection{The Pseudo-code of TISVD}
\label{TISVD code}
In this section, we provide the pseudo-code of TISVD.

\begin{algorithm}[htp]
    \caption{{\sf TISVD}} \label{alg:ISVD}
    \textbf{Input}: Rank-$k$ singular matrix $\bm{U}^{(t)}$, $\bm{V}^{(t)}$ and 
    $\boldsymbol{\Sigma}^{(t)}$ at round $t$, low-rank matrix $\bm{\Delta}_1$ and $\bm{\Delta}_2$,
    truncated rank $k$ \\
    \textbf{Output}: Rank-$k$ singular matrix $\bm{U}^{(t+1)}$, $\bm{V}^{(t+1)}$ and 
    $\boldsymbol{\Sigma}^{(t+1)}$ \\
    \begin{algorithmic}[1]
    \STATE $\bm{U}_A \gets \Big(\bm{I}-\bm{U}^{(t)}\bm{U}^{(t)\top}\Big)\bm{\Delta}_1$
    \STATE $\bm{V}_B \gets \Big(\bm{I}-\bm{V}^{(t)}\bm{V}^{(t)\top}\Big)\bm{\Delta}_2$ 
    \STATE Compute orthogonal basis $\bm{P},\bm{Q}$ of the column space of $\bm{U}_A,\bm{V}_B$, respectively. 
    \STATE Set $\bm{R}_1 \gets \bm{P}^{\top}\Big(\bm{I}-\bm{U}^{(t)}\bm{U}^{(t)\top}\Big)\bm{\Delta}_1$
    \STATE Set $\bm{R}_2 \gets \bm{Q}^{\top}\Big(\bm{I}-\bm{V}^{(t)}\bm{V}^{(t)\top}\Big)\bm{\Delta}_2$
    \STATE $\bm{H} \gets \begin{bmatrix}
    \boldsymbol{\Sigma}^{(t)}& \bm{0} \\
    \bm{0} & \bm{0}
    \end{bmatrix} + 
    \begin{bmatrix}
    \bm{U}^{(t)\top}\bm{\Delta}_1 \\
    \bm{R}_1
    \end{bmatrix}
    \begin{bmatrix}
    \bm{V}^{(t)\top}\bm{\Delta}_2 \\
    \bm{R}_2
    \end{bmatrix}^{\top}
    $
    \STATE Compute $\tilde{\bm{U}}_k$, $\tilde{\bm{V}}_k$ and $\tilde{\boldsymbol{\Sigma}}_k$ from rank-$k$ SVD of $\bm{H}$ \;
    \STATE \# Update singular matrix. 
    \STATE $\bm{U}^{(t+1)} \gets \begin{bmatrix}
    \bm{U}^{(t)} & P 
    \end{bmatrix}\tilde{\bm{U}} $ 
    \STATE $\bm{V}^{(t+1)} \gets \begin{bmatrix}
    \bm{V}^{(t)} & Q 
    \end{bmatrix}\tilde{\bm{V}} $ 
    \STATE $\boldsymbol{\Sigma}^{(t+1)} \gets \tilde{\boldsymbol{\Sigma}}$ 
    \STATE \textbf{return} $\bm{U}^{(t+1)}, \bm{V}^{(t+1)}, \boldsymbol{\Sigma}^{(t+1)}$
    \end{algorithmic}
\end{algorithm}

\subsection{More Discussion about TISVD}
\label{TISVD Discussion}

Current incremental SVD methods necessitate the prerequisite that the decomposition matrix adheres to a low-rank structure~\cite{brand2006fast}. 
When this low-rank condition isn't met, these methods devolve into traditional SVD. 
However, in online learning scenarios, the assurance of a low-rank decomposed sketch matrix isn't guaranteed. 
In this context, TISVD innovatively accomplishes incremental maintenance of singular value matrices without relying on low-rank assumptions, rendering it adapt to online learning algorithms founded on incremental sketching methodologies.

More precisely, given the matrix $\bm{A}=\bm{U}\boldsymbol{\Sigma}\bm{V}^{\top}\in\mathbb{R}^{n\times{n}}$, the conventional incremental SVD (ISVD) streamlines the process by omitting the rotation and re-orthogonalization of $\bm{U}$ and $\bm{V}$, leading to a time complexity of $O(nr+r^3)$. where $r$ denotes the matrix rank. 
Consequently, ISVD relies on the assumption that $r\ll{n}$ in order to effectively establish a linear-time SVD algorithm.

Nevertheless, in online learning scenarios, the sketch matrix earmarked for decomposition frequently fails to adhere to the low-rank characteristic, thereby rendering the direct application of ISVD ineffective in achieving linear time complexity. To counter this predicament, we have integrated truncation techniques within the framework of traditional incremental SVD methods. This adaptation yields a time complexity of $O(nr_t+r_t^3)$, with $r_t$ signifying the predetermined truncated rank. Crucially, this truncation innovation positions TISVD as a linear incremental SVD technique that stands independent of low-rank assumptions.

As previously discussed, applying ISVD directly to online learning algorithms isn't viable. 
Recognizing the substantial enhancement that accelerated decomposition feature mapping can offer to the performance of online kernel learning algorithms, prevailing research employs the randomized SVD algorithm to expedite these algorithms~\cite{wan2021efficient,zhang2019incremental}. 
However, it's important to highlight that, unlike the incremental SVD method, the randomized SVD is a rapid SVD technique reliant on random matrices and lacks the capability to perform incremental updates on singular value matrices.
Furthermore, it's worth noting that the time complexity of randomized SVD is $O(n^2r+r^3)$, which is comparatively slower than TISVD's $O(nr+r^3)$.

We have compared TISVD with the rank-$k$ truncated SVD in Section 5.2. We further construct an experiment to test the performance of randomized SVD. We initialize a random Gaussian matrix $\bm{A}\in\mathbb{R}^{100\times100}$ and use random Gaussian matrix $B, C\in\mathbb{R}^{m\times100}$ as low rank update. We set $m=3,k=30$ and update $\bm{A}$ $500$ to $50000$ times respectively.

\begin{figure}[htp]
    \centering
    \includegraphics[width=8.5cm]{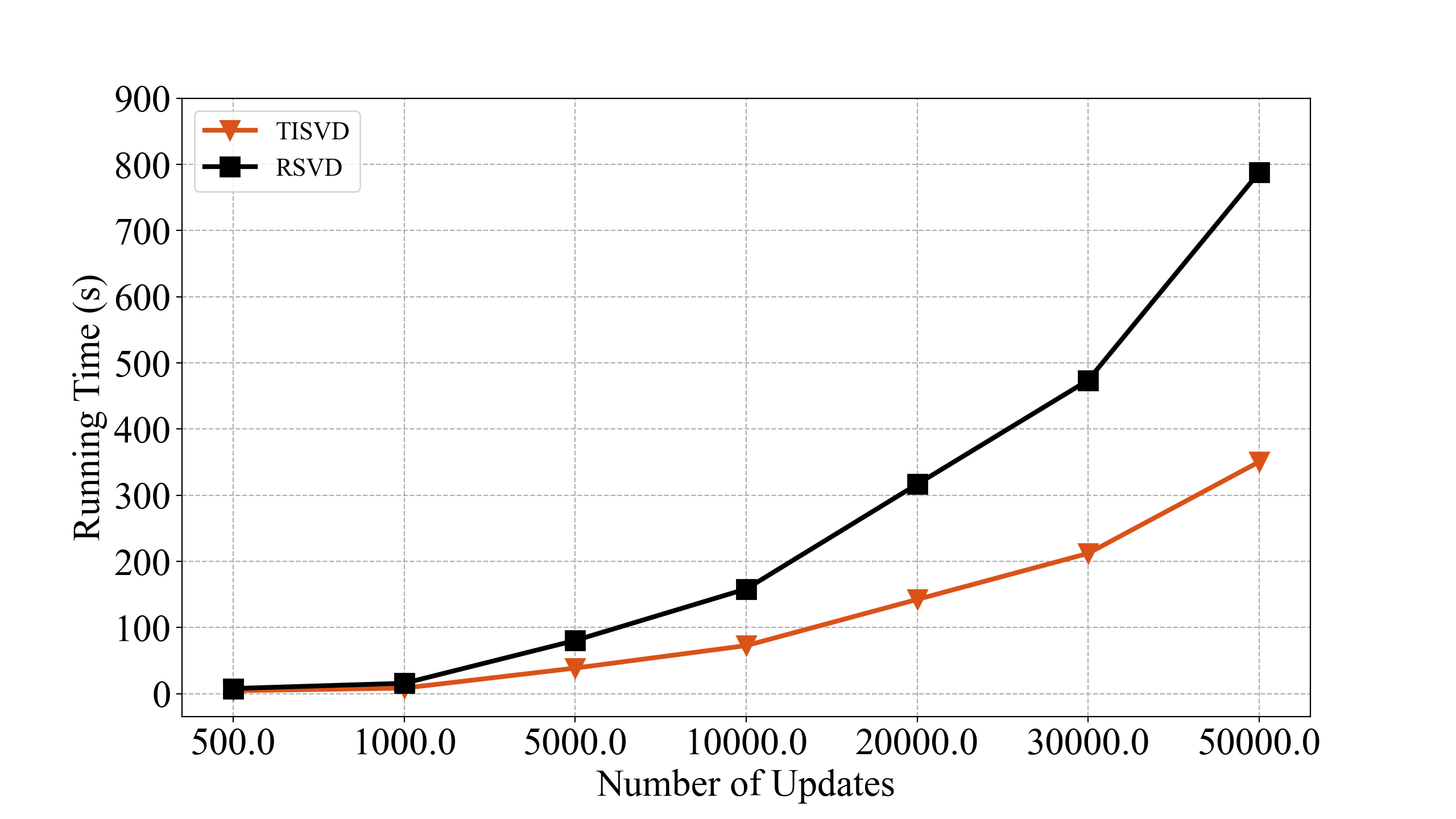}
    \caption{The comparison of running time between TISVD and RSVD}
    \label{fig: RSVD exp}
\end{figure}

From Figure~\ref{fig: RSVD exp}, we see that TISVD continues to show desirable decomposition performance.
Both TIVD and randomized SVD can reduce the size of the decomposed matrix through the low-rank approximation matrix, thereby accelerating the algorithm. 
However, randomized SVD enjoys a time complexity of $O(n^2k+k^3)$, which is worse than $O(nk+k^3)$.
Besides, TISVD uses incremental updates to update the singular value matrix, which is more scalable for online learning algorithms based on incremental sketching.

\section{Omitted Proofs for Section~\ref{sec: Regret Analysis}}

In this section, we present the omitted proofs from Section~\ref{sec: Regret Analysis}. 
The proof of Theorem~\ref{thm:OnlineKernel:Sketch:SJLT:regret} is provided in Appendix~\ref{sec: proof of Thm1}. 
Additionally, we prove the worst-case regret when Assumption~\ref{assm:OKL:Curvature} is relaxed to the convex case in Appendix~\ref{Remark 2 proof}.

\subsection{Proof of Theorem~\ref{thm:OnlineKernel:Sketch:SJLT:regret}}
\label{sec: proof of Thm1}

We can refine the representation of the difference in losses between $f^*$ and $\bm w^*$ using the approximation error of the kernel matrix, 
where $f^*$ is the optimal hypothesis in the original RKHS in hindsight, and $\bm w^*$ is the optimal hypothesis on the incremental randomized sketches in hindsight.
Specifically, we utilize the following conclusion from Theorem 2 in \cite{yang2012nystrom}:
    \begin{equation*}
        \ell (\bm w^*) - \ell (f^*) \leq \dfrac{1}{2T\lambda}
        \| {\bm \Tilde{\bm{K}}  }_{\mathrm{sk}}^{(T)} - \bm K  \|_2,
    \end{equation*}
    
    yielding that
    \begin{equation}
    \label{eq:IRSS:regret:hindsight}
    \begin{aligned}
        \sum_{t=1}^{T} \left( \ell_t  (\bm w^*) - \ell_t (f^*) \right) 
        &\leq
        \dfrac{1}{2\lambda}
             \left\|
                [{\bm \Tilde{\bm{K}}  }_{\mathrm{sk}}^{(T)}]_{B, \rho}-[{\bm K  }_{\mathrm{sk}}^{(T)}]_{B, \rho}
             \right\|_2+ \\
        &\hphantom{{}={}} \dfrac{1}{2\lambda}
             \left\|
                [{\bm K  }_{\mathrm{sk}}^{(T)}]_{B, \rho} - {\bm K}_{B,\rho}
             \right\|_2 + \\ 
             &\hphantom{{}={}}
             \dfrac{1}{2\lambda}\left\|
               \widehat{\bm K}_{B,\rho} - \bm K
             \right\|_2
    \end{aligned}
    \end{equation}
where $\bm K_{B,\rho} \in \mathbb{R}^{(B + \lfloor (T-B) / \rho \rfloor)\times (B + \lfloor (T-B) / \rho \rfloor)}$ is the intersection matrix of $\bm K$, constructed using $B + \lfloor (T-B) / \rho \rfloor$ examples, $[{\bm \Tilde{\bm{K}} }_{\mathrm{sk}}^{(T)}]_{B, \rho}$ is the approximate matrix for $\bm K_{B,\rho}$ obtained using the proposed incremental sketching method and TISVD with rank parameter $k$, $[{\bm K }_{\mathrm{sk}}^{(T)}]_{B, \rho}$ is the best rank $k$ approximate matrix for $\bm K_{B,\rho}$ obtained using the proposed incremental sketching method, $\bm O$ is a zero matrix of size $(T - B - \lfloor (T-B) / \rho \rfloor) \times (T - B - \lfloor (T-B) / \rho \rfloor)$, and
    \begin{align*}
        \widehat{\bm K}_{B,\rho} & =
        \mathrm{diag}
        \left\{  \bm K_{B,\rho},~ \bm O \right\}
        \in \mathbb{R}^{T \times T}.
    \end{align*}

    TISVD truncates to rank $k$ at each update round, leading to an elevated spectral loss as the number of update rounds increases. Nevertheless, the polynomial decay property of the kernel matrix and our periodic updating strategy enable effective control of this loss.
    Given that the eigenvalues of the kernel matrix decay polynomially with a decay rate $\beta > 1$, we have:
    \begin{equation}
        \label{eq:TISVD:matrix}
        \begin{aligned}
             \left\|
                [{\bm \Tilde{\bm{K}}  }_{\mathrm{sk}}^{(T)}]_{B, \rho}-[{\bm K  }_{\mathrm{sk}}^{(T)}]_{B, \rho}
             \right\|_2 \le \frac{1}{\theta}(k+1)^{-\beta}
        \end{aligned}
    \end{equation}
    
    Since the eigenvalues of the kernel matrix decay polynomially, we can establish the following bound:
    \begin{equation}
    \label{eq:IRSS:regret:matrix:decay}
    \begin{aligned}
        \|{\widehat{\bm K} }_{B,\rho}  - \bm K   \|_2
        &\leq
        \dfrac{T - B - \lfloor (T-B) / \rho \rfloor}{T}
        \sum_{i=1}^{T} i^{-\beta}
        \\
        &\leq
        \dfrac{T - B - \lfloor (T-B) / \rho \rfloor}{T}
        \int_{1}^{T} i^{-\beta} \mathrm{d} i
        \\
        &=
        \dfrac{T - B - \lfloor (T-B) / \rho \rfloor}{T}
        \dfrac{1}{\beta-1} \left( 1 - \dfrac{1}{T^{\beta - 1}}\right)
        \\
        &\leq
        \dfrac{1}{\beta-1}
        \left(1 -  \dfrac{B+ \lfloor (T-B)/\rho \rfloor}{T}\right).
    \end{aligned}
    \end{equation}
    Besides, from Assumption 3,
    with probability at least $1 - \delta$,
    we have 
    \begin{equation}
    \label{eq:IRSS:regret:matrix:approximation}
    \begin{aligned}
        &\hphantom{{}={}}
         \left\| [{\bm K  }_{\mathrm{sk}}^{(T)}]_{B,\rho} -  \bm K_{B,\rho} \right\|_2& \\
        &\leq
          \sqrt{ 1 + \epsilon}~
          \| [\bm C_{\mathrm{m}} \bm F_{\mathrm{mod}} {\bm C}_{\mathrm{m}}^{\top}]_{B,\rho}- \bm K_{B,\rho} \|_{\mathrm{F}},
    \end{aligned}
    \end{equation}
    where $[\bm C_{\mathrm{m}} \bm F_{\mathrm{mod}} {\bm C}_{\mathrm{m}}^{\top}]_{B,\rho}$ is the approximate matrix for $\bm K_{B,\rho}$ using the modified Nyström approach with a rank parameter $k$.

    Denoting the best rank-$k$ approximation of $\bm A$ as $(\bm A)_k$, and considering that the eigenvalues of $\bm K$ decay polynomially with a decay rate $\beta > 1$, we can find a value of $\beta > 1$ such that $\lambda_i(\bm K) = O(i^{-\beta})$. This leads to the following expression:
    \begin{equation}
    \label{eq:IRSS:rank_k}
    \begin{aligned}
       &\hphantom{{}={}}
       \| \bm K_{B,\rho} - (\bm K_{B,\rho})_k \|_{\mathrm{F}} &
       \\
       &=\sqrt{B+ \lfloor (T-B)/\rho \rfloor - k} \cdot (k+1)^{-\beta} \\
       &= O(\sqrt{B}).
    \end{aligned}
    \end{equation}
    Given $\epsilon' \in (0, 1)$, when $s_{\mathrm{m}} = \Omega ( \mu(\bm K_{B,\rho}) k \log k)$, according to Theorem 22 in~\cite{wang2016spsd}, we can derive the following bound:
    \begin{equation}\label{eq:IRSS:regret:uniform:bound}
    \begin{aligned}
        &\hphantom{{}={}}
        \left\| [\bm C_{\mathrm{m}} \bm F_{\mathrm{mod}} {\bm C}_{\mathrm{m}}^{\top}]_{B,\rho}- \bm K_{B,\rho} \right\|_{\mathrm{F}}\\
        &\leq
        \left\|
            [\bm C_{\mathrm{m}} \bm F_{\mathrm{mod}} {\bm C}_{\mathrm{m}}^{\top}]_{B,\rho} -
            [\bm C_{\mathrm{m}} \bm C_{\mathrm{m}}^\dagger \bm K_{B,\rho}]_{B,\rho}
        \right\|_{\mathrm{F}}+\\
        &\hphantom{{}={}}
        \left\| [\bm C_{\mathrm{m}} \bm C_{\mathrm{m}}^\dagger \bm K_{B,\rho}]_{B,\rho} - \bm K_{B,\rho} \right\|_{\mathrm{F}} \\
        &=
        \left\|
            [\bm C_{\mathrm{m}} \bm C_{\mathrm{m}}^{\dagger}
            {\bm K}_{B,\rho}
            \left( \bm C_{\mathrm{m}} \bm C_{\mathrm{m}}^{\dagger}  \right)^{\top} -
            \bm C_{\mathrm{m}} \bm C_{\mathrm{m}}^\dagger \bm K_{B,\rho}]_{B,\rho}
        \right\|_{\mathrm{F}} +\\
        &\hphantom{{}={}}
        \left\| [\bm C_{\mathrm{m}} \bm C_{\mathrm{m}}^\dagger \bm K_{B,\rho}]_{B,\rho} - \bm K_{B,\rho} \right\|_{\mathrm{F}}
        \\
        &\leq
        \left\|
            [\bm C_{\mathrm{m}} \bm C_{\mathrm{m}}^{\dagger}]_{B,\rho}
        \right\|_{\mathrm{F}}
        \left\|
            [
            {\bm K}_{B,\rho}
            \left( \bm C_{\mathrm{m}} \bm C_{\mathrm{m}}^{\dagger}  \right)^{\top}
            ]_{B,\rho} -
            \bm K_{B,\rho}
        \right\|_{\mathrm{F}} +\\
        &\hphantom{{}={}}
        \left\| [\bm C_{\mathrm{m}} \bm C_{\mathrm{m}}^\dagger \bm K_{B,\rho}]_{B,\rho} - \bm K_{B,\rho} \right\|_{\mathrm{F}}
        \\
        &=
        \left(
            1 +
            \left\|
                [\bm C_{\mathrm{m}} \bm C_{\mathrm{m}}^{\dagger}]_{B,\rho}
            \right\|_{\mathrm{F}}
        \right)
        \left\| [\bm C_{\mathrm{m}} \bm C_{\mathrm{m}}^\dagger \bm K_{B,\rho}]_{B,\rho} - \bm K_{B,\rho} \right\|_{\mathrm{F}}
        \\
        &\leq
        (1 + \sqrt{s_{\mathrm{m}}})
        \left\| [\bm C_{\mathrm{m}} \bm C_{\mathrm{m}}^\dagger \bm K_{B,\rho}]_{B,\rho} - \bm K_{B,\rho} \right\|_{\mathrm{F}}
        \\
        &\leq \sqrt{1 + \epsilon'}~
        (1 + \sqrt{s_{\mathrm{m}}})
        \| \bm K_{B,\rho} - (\bm K_{B,\rho})_k \|_{F},
    \end{aligned}
    \end{equation} 
where $[\bm A]_{B,\rho}$ indicates that $\bm A$ is constructed based on the matrix $\bm K_{B,\rho}$, and $\mu(\bm K_{B,\rho})$ represents the coherence of $\bm K_{B,\rho}$. By combining \eqref{eq:IRSS:regret:matrix:approximation}, \eqref{eq:IRSS:rank_k}, and \eqref{eq:IRSS:regret:uniform:bound}, we obtain the following result:   \begin{equation}\label{eq:IRSS:regret:matrix:approximation:final}
         \left\| [{\bm K  }_{\mathrm{sk}}^{(T)}]_{B,\rho} -  \bm K_{B,\rho} \right\|_2
          \leq
          \sqrt{ 1 + \epsilon}~
          O(\sqrt{B}).
    \end{equation}
    Substituting \eqref{eq:TISVD:matrix}, \eqref{eq:IRSS:regret:matrix:decay} and
    \eqref{eq:IRSS:regret:matrix:approximation:final}
    into \eqref{eq:IRSS:regret:hindsight},
    we have
    \begin{equation}\label{eq:IRSS:regret:matrix:approximation:substituting}
    \begin{aligned}
        &\hphantom{{}={}}
        \sum_{t=1}^{T} \left( \ell_t  (\bm w^*) - \ell_t (f^*) \right) \\
        &\leq
             \dfrac{1}{2\lambda(\beta-1)}
             \left(1 -  \dfrac{B+ \lfloor (T-B)/\rho \rfloor}{T}\right)+
              \dfrac{\sqrt{ 1 + \epsilon}}{2\lambda}O(\sqrt{B}) \\
             &\quad
              + \dfrac{(k+1)^{-\beta}}{2\lambda}\lfloor \frac{T-B}{\rho}\rfloor
              + \frac{\lambda}{2} \| f^* \|_{\mathcal{H}_{\kappa}}^2 - \frac{\lambda}{2} \| \bm w^* \|_2^2.
    \end{aligned}
    \end{equation}
    Next, we analyze the regret resulting from hypothesis updating on the incremental randomized sketches. We begin by decomposing $ \ell_t (\bm w_t) - \ell_t (\bm w^*)$ into two terms as follows:
    \begin{align*}
        \ell_t (\bm w_t) - \ell_t (\bm w^*)
        =
        \underbrace{\ell_t (\bm w_t) - \ell_t (\bm w_t^*)}_{\text{Term 1: Optimization Error}} +
        \underbrace{\ell_t (\bm w_t^*) - \ell_t (\bm w^*)}_{\text{Term 2: Estimation Error}},
    \end{align*}
    where $f^*_t (\cdot)= \langle \bm w_t^*, \bm \phi_t(\cdot)\rangle$ represents the optimal hypothesis on the incremental sketches for the first $t$ instances, and $\bm w^*$ denotes the optimal hypothesis on the incremental sketches in hindsight. 
    
    The optimization error quantifies the discrepancy between the hypothesis generated by the proposed faster second-order online kernel learning algorithm and the optimal hypothesis on the incremental randomized sketches at each round. On the other hand, the estimation error measures the difference between the optimal hypotheses on the incremental randomized sketches for the first $t$ instances and for all $T$ instances, respectively.

    To obtain an upper bound for the optimization error, we leverage the directional curvature condition presented in Assumption 2. Given that the Euclidean regularization is a strongly convex regularizer, the loss function $\ell_t$ also satisfies the directional curvature condition. As a result, we can utilize the inequality provided in Assumption 2 to bound the optimization error. 
    Specifically, we obtain the following expression:
    \begin{equation}\label{eq:FORKS:regret:term1}
    \begin{aligned}
    &\hphantom{{}={}}
    \ell_t (\bm w_t) - \ell_t (\bm w_t^*)\\
    &\leq
    \left\langle \nabla \ell_t ( \bm w_t), \bm w_t - \bm w_t^* \right\rangle-
    \dfrac{L_{\mathrm{Cur}}}{2}
    \left\langle \nabla \ell_t ( \bm w_t), \bm w_t^* - \bm w_t \right\rangle^2.
    \end{aligned}
    \end{equation}
    Letting
    $$
        \Delta_t = \left\langle \nabla \ell_t ( \bm w_t), \bm w_t - \bm w_t^* \right\rangle -
    \dfrac{L_{\mathrm{Cur}}}{2}
    \left\langle \nabla \ell_t ( \bm w_t), \bm w_t^* - \bm w_t \right\rangle^2,
    $$
    \eqref{eq:FORKS:regret:term1} can be rewritten as $\ell_t (\bm w_t) - \ell_t (\bm w_t^*)
    \leq \Delta_t.$
    Note that $\bm g_t = \nabla \ell_t ( \bm w_t)$ in the FORKS algorithm, we first give the bound of $ \left\langle \bm g_t, \bm w_t - \bm w_t^* \right\rangle = \left\langle \nabla \ell_t ( \bm w_t), \bm w_t - \bm w_t^* \right\rangle $ in $\Delta_t$.
    Based on the update steps for $\bm v_t$ and $\bm w_t$ proposed in FORKS, it can be inferred that
    \begin{equation*}
    \begin{aligned}
        \bm v_{t+1} - \bm w_t^* &= \bm w_t - \bm w_t^* - \bm A_t^{-1} \bm g_t, \\
        \bm A_t (\bm v_{t+1} - \bm w_t^* ) &= \bm A_t (\bm w_t - \bm w_t^*) - \bm g_t,
    \end{aligned}
    \end{equation*}
    yielding that
   \begin{equation}\label{eq:FORKS:regret:vw}
    \begin{aligned}
        &\hphantom{{}={}} \langle \bm v_{t+1} - \bm w_t^*, \bm A_t (\bm v_{t+1} - \bm w_t^* ) \rangle \\
        &=
        \langle \bm w_t - \bm w_t^*, \bm A_t (\bm w_t - \bm w_t^*) \rangle - 2\langle \bm g_t, \bm w_t - \bm w_t^*  \rangle +\\
        &\hphantom{{}={}}
        \langle \bm g_t, \bm A_t^{-1} \bm g_t\rangle.
    \end{aligned}
    \end{equation}
    Considering that $\bm w_{t+1}$ in FORKS can be interpreted as the generalized projection of $\bm v_{t+1}$ within the norm induced by $\bm A_t$, by leveraging \eqref{eq:FORKS:regret:vw} and the Pythagorean theorem, we can derive the following relationship:
   \begin{equation}\label{eq:FORKS:regret:vw:two}
    \begin{aligned}
        &\hphantom{{}={}} 2\langle \bm g_t, \bm w_t - \bm w_t^* \rangle \\
        &=
        \langle \bm w_t - \bm w_t^*, \bm A_t (\bm w_t - \bm w_t^*) \rangle +
        \langle \bm g_t, \bm A_t^{-1} \bm g_t\rangle -\\
        &\hphantom{{}={}}
        \langle \bm v_{t+1} - \bm w_t^*, \bm A_t (\bm v_{t+1} - \bm w_t^* ) \rangle
        \\
        &\leq
        \langle \bm w_t - \bm w_t^*, \bm A_t (\bm w_t - \bm w_t^*) \rangle +
       \langle \bm g_t, \bm A_t^{-1} \bm g_t\rangle -\\
       &\hphantom{{}={}}
        \langle \bm w_{t+1} - \bm w_t^*, \bm A_t (\bm w_{t+1} - \bm w_t^* ) \rangle        .
    \end{aligned}
    \end{equation}
    By summing \eqref{eq:FORKS:regret:vw:two} for $t \in [T]$, combining with \eqref{eq:FORKS:regret:term1} we obtain
    \begin{equation}\label{eq:FORKS:regret:vw:two:sum}
    \begin{aligned}
    &\hphantom{{}={}}
    \sum_{t=1}^{T} \ell_t (\bm w_t) - \ell_t (\bm w_t^*)\\
    &\leq
    \sum_{t=1}^{T}   \langle \bm g_t, \bm w_t - \bm w_t^* \rangle
    -
    \sum_{t=1}^{T}
    \dfrac{L_{\mathrm{Cur}}}{2} \left\langle \nabla \ell_t ( \bm w_t), \bm w_t^* - \bm w_t \right\rangle^2
    \\
        &\leq
        \frac{1}{2}\sum_{t=1}^{T} \langle \bm w_t - \bm w_t^*, \bm A_t (\bm w_t - \bm w_t^*) \rangle +
        \frac{1}{2}\sum_{t=1}^{T} \langle \bm g_t, \bm A_t^{-1} \bm g_t\rangle -
        \\
        &\hphantom{{}={}}
        \frac{1}{2}\sum_{t=1}^{T} \langle \bm w_{t+1} - \bm w_t^*, \bm A_t (\bm w_{t+1} - \bm w_t^* ) \rangle
        - \\
        &\hphantom{{}={}}
        \sum_{t=1}^{T}
        \dfrac{L_{\mathrm{Cur}}}{2} \left\langle \nabla \ell_t ( \bm w_t), \bm w_t^* - \bm w_t \right\rangle^2.
    \end{aligned}
    \end{equation}
    Since incremental sketches in FORKS are periodically updated, $\bm w_t^*$ can be updated at most $\lfloor (T-B) /\rho \rfloor$ times. Consequently, by leveraging the fact that $\bm{A}_{t+1} = \bm{A}_{t}+ \sigma_t \bm{g}_t \bm g_t^\top$, where $\sigma_t\ge L_{\mathrm{Cur}}$, the upper bound in~\eqref{eq:FORKS:regret:vw:two:sum} can be simplified to the following expression:
    \begin{equation}\label{eq:FORKS:regret:vw:two:sum:displace}
    \begin{aligned}
    &\hphantom{{}={}}
    \sum_{t=1}^{T} \ell_t (\bm w_t) - \ell_t (\bm w_t^*)
    \\
        &\leq
        \frac{1}{2} \langle \bm w_1 - \bm w_1^*, (\bm A_2 - \frac{\bm g_1 \bm g_1^\top}{2} ) (\bm w_1 - \bm w_1^*) \rangle+
        \\
        &\hphantom{{}={}}
        \frac{1}{2} \sum_{t=1}^T(\bm{w}_t-\bm{w}_t^*)^\top (\bm{A}_t-\bm{A}_{t-1}-\sigma_t\bm{g}_t\bm{g}_t^\top)(\bm{w}_t-\bm{w}_t^*)+
        \\
        &\hphantom{{}={}}
        \frac{1}{2} \sum_{t=1}^{T} \langle \bm g_t, \bm A_t^{-1} \bm g_t\rangle
        \\
        &=
        \frac{1}{2}  \langle \bm w_1 - \bm w_1^*, \bm A_1  (\bm w_1 - \bm w_1^*) \rangle +
        \frac{1}{2}\sum_{t=1}^{T} \langle \bm g_t, \bm A_t^{-1} \bm g_t\rangle + \\
        &\hphantom{{}={}} 
        \sum_{t=1}^T\frac{\eta_t}{2}(\bm{w}_t-\bm{w}_t^*)^\top \bm{g}_t\bm{g}_t^\top(\bm{w}_t-\bm{w}_t^*) 
        \\
        &=
        \frac{\alpha}{2} \left\| \bm w_1 - \bm w_1^* \right\|_2^2 +
        \frac{1}{2}\sum_{t=1}^{T} \langle \bm g_t, \bm A_t^{-1} \bm g_t\rangle
        \\
        &\leq
        \frac{\alpha D_{\bm w}^2}{2}  +
        \frac{1}{2}\sum_{t=1}^{T} \langle \bm g_t, \bm A_t^{-1} \bm g_t\rangle,
    \end{aligned}
    \end{equation}
    By applying the result from~\cite{hazan2007logarithmic}, we can obtain the following upper bound on the sum of the inner products $\langle \bm g_t, \bm A_t^{-1} \bm g_t\rangle, \forall t \in [T]$:
    \begin{equation}\label{eq:FORKS:regret:vw:two:sum:displace:gAg}
    \begin{aligned}
         \sum_{t=1}^{T} \langle \bm g_t, \bm A_t^{-1} \bm g_t\rangle 
         &\leq 
         \frac{1}{L_{\mathrm{Cur}}}\log\left(T L_{\mathrm{Lip}}^2/L_{\mathrm{Cur}} + 1 \right)^k \\
         &= 
         \frac{k}{L_{\mathrm{Cur}}} \log\left(T L_{\mathrm{Lip}}^2/L_{\mathrm{Cur}} + 1 \right).
    \end{aligned}
    \end{equation}
    Combining \eqref{eq:FORKS:regret:vw:two:sum:displace:gAg} with \eqref{eq:FORKS:regret:vw:two:sum:displace},
    we find that
    \begin{align}\label{eq:IRSS:regret:term1}
      &\hphantom{{}={}}
      \sum_{t=1}^{T} \left( \ell_t (\bm w_t) - \ell_t (\bm w_t^*) \right) \nonumber\\
      &\leq
        \frac{\alpha D_{\bm w}^2}{2}  +
        \frac{k}{2L_{\mathrm{Cur}}}  O(\log T) +
        \frac{\lambda}{2} \| \bm w_t^* \|_2^2 - \frac{\lambda}{2} \| \bm w_t \|_2^2.
    \end{align}

    For the estimation error,
    we obtain the following upper bound
    \begin{equation}
    \label{eq:IRSS:regret:term2}
    \begin{aligned}
    &\hphantom{{}={}}
      \sum_{t=1}^{T} \left( \ell_t (\bm w_t^*) - \ell_t (\bm w^*) \right)\\
      &\leq
      \dfrac{1}{2\lambda} \left\| {\bm K  }_{\mathrm{sk}}^{(T_0)} - {\bm K  }_{\mathrm{sk}}^{(T)} \right\|_2
      + \frac{\lambda}{2} \| \bm w^* \|_2^2 - \frac{\lambda}{2} \| \bm w_t^* \|_2^2
      \\
       &\leq
       \dfrac{1}{2\lambda}
       \left(
            \left\| {\bm K  }_{\mathrm{sk}}^{(T_0)} - \bm K^{(T_0)}\right\|_2 +
            \left\| \bm K^{(T_0)} - {\bm K} \right\|_2
       \right) +\\
       &\hphantom{{}={}} 
       \dfrac{1}{2\lambda}\left\| {\bm K  }_{\mathrm{sk}}^{(T)} - {\bm K  }  \right\|_2+\frac{\lambda}{2} \| \bm w^* \|_2^2- \frac{\lambda}{2} \| \bm w_t^* \|_2^2
       \\
       &\leq
       \dfrac{1}{2\lambda}
       \left[
                \sqrt{ 1 + \tilde{\epsilon}}~O(\sqrt{B})+
               \dfrac{1}{\beta-1}
                \left(1 -  \dfrac{B}{T}\right)
       \right]
       +\\
       &\hphantom{{}={}}
       \dfrac{1}{2\lambda}\left\| {\bm K  }_{\mathrm{sk}}^{(T)} - {\bm K    }  \right\|_2+
       \frac{\lambda}{2} \| \bm w^* \|_2^2 - \frac{\lambda}{2} \| \bm w_t^* \|_2^2.
    \end{aligned}
    \end{equation}
    Finally,
    the three inequalities \eqref{eq:IRSS:regret:matrix:approximation:substituting}, \eqref{eq:IRSS:regret:term1}
    and  \eqref{eq:IRSS:regret:term2} combined give the following bound:
    \begin{align*}
        &\hphantom{{}={}}
        \sum_{t=1}^{T}\left( \ell_t (\bm w_t) - \ell_t (f^*) \right)\\
        &\leq
       \frac{\alpha D_{\bm w}^2}{2}  +
        \frac{k}{2L_{\mathrm{Cur}}}  O(\log T) + \dfrac{(k+1)^{-\beta}}{2\lambda\theta}+\\
        &\hphantom{{}={}}
        \frac{\lambda}{2} \| f^* \|_{\mathcal{H}_{\kappa}}^2+
        \dfrac{1}{\lambda(\beta-1)}
        \left( \dfrac{3}{2} -  \dfrac{B+ \lfloor (T-B)/\rho \rfloor}{T}\right)+ \\
        &\hphantom{{}={}}
        \dfrac{\sqrt{1 + \epsilon}}{\lambda}O(\sqrt{B}).
    \end{align*}

\subsection{Proof of Convex Case}
\label{Remark 2 proof}
By leveraging the fact that $\bm{A}_{t+1} = \bm{A}_{t}+ (\sigma_t+\eta_t) \bm{g}_t \bm g_t^\top$, where $\sigma_t\ge L_{\mathrm{Cur}}$ and applying the Proposition 1 from \cite{luo2016efficient}, we can rewrite \eqref{eq:FORKS:regret:vw:two:sum:displace} as:
\begin{equation}
\label{eq:FORKS:regret:vw:two:sum:displace2}
    \begin{aligned}
    &\hphantom{{}={}}
    \sum_{t=1}^{T} \ell_t (\bm w_t) - \ell_t (\bm w_t^*)
    \\
        &\leq
        \frac{1}{2}  \langle \bm w_1 - \bm w_1^*, \bm A_1  (\bm w_1 - \bm w_1^*) \rangle +
        \frac{1}{2}\sum_{t=1}^{T} \langle \bm g_t, \bm A_t^{-1} \bm g_t\rangle + \\
        &\hphantom{{}={}}
        \sum_{t=1}^T\frac{\eta_t}{2}(\bm{w}_t-\bm{w}_t^*)^\top \bm{g}_t\bm{g}_t^\top(\bm{w}_t-\bm{w}_t^*) 
        \\
        &=
        \frac{\alpha}{2} \left\| \bm w_1 - \bm w_1^* \right\|_2^2 +
        \frac{1}{2}\sum_{t=1}^{T} \langle \bm g_t, \bm A_t^{-1} \bm g_t\rangle +
        \\
        &\hphantom{{}={}}
        \sum_{t=1}^T\frac{\eta_t}{2}(\bm{w}_t-\bm{w}_t^*)^\top \bm{g}_t\bm{g}_t^\top(\bm{w}_t-\bm{w}_t^*)
        \\
        &\leq
        \frac{\alpha D_{\bm w}^2}{2}  +
        \frac{1}{2}\sum_{t=1}^{T} \langle \bm g_t, \bm A_t^{-1} \bm g_t\rangle +
        2L_{\mathrm{Lip}}^2\sum_{t=1}^T\eta_t
        \\
        &\leq
        \frac{\alpha D_{\bm w}^2}{2}  +
        \frac{k}{2(\eta_T+L_{\mathrm{Cur}})}  O(\log T) +
        2L_{\mathrm{Lip}}^2\sum_{t=1}^T\eta_t,
    \end{aligned}
    \end{equation}

In the worst case, if $L_{\mathrm{Cur}} = 0$, we set $\eta_t = \sqrt{\frac{k}{L_{\mathrm{Lip}}^2t}}$ and the bound can be simplified to:

\begin{align*}
        &\hphantom{{}={}}
        \sum_{t=1}^{T}\left( \ell_t (\bm w_t) - \ell_t (f^*) \right)\\
        &\leq
       \frac{\alpha D_{\bm w}^2}{2}  +
        \frac{\sqrt{k}L_{\mathrm{Lip}}}{2}  O\left(\sqrt{T}\right) + 4\sqrt{k}L_{\mathrm{Lip}}O\left(\sqrt{T}\right) + 
        \\
       &\hphantom{{}={}}
       \frac{\lambda}{2} \| f^* \|_{\mathcal{H}_{\kappa}}^2 +
        \dfrac{1}{\lambda(\beta-1)}
        \left( \dfrac{3}{2} -  \dfrac{B+ \lfloor (T-B)/\rho \rfloor}{T}\right)
        \\
       &\hphantom{{}={}}
       +
        \dfrac{\sqrt{1 + \epsilon}}{\lambda}O(\sqrt{B}) + \dfrac{(k+1)^{-\beta}}{2\theta\lambda}.
\end{align*}


\section{Omitted Details for Section~\ref{sec: Experiments}}

In this section, we provide the missing details and additional experimental results from Section~\ref{sec: Experiments}. 
Appendix~\ref{Dataset and Experimental Setup} contains detailed information on the dataset and experimental setup. The experimental settings for the KuaiRec dataset are outlined in Appendix~\ref{Apx: KuaiRec}. 
Additional experimental results under adversarial conditions and with large-scale real-world datasets are presented in Appendix~\ref{sec: Additional Experiment Results under Adversarial Environment} and Appendix~\ref{sec: Additional Experiment Results under Large-Scale Real-world Datasets}, respectively.

\subsection{Dataset and Experimental Setup}
\label{Dataset and Experimental Setup}
We evaluate FORKS on several real-world datasets for binary classification tasks. 
We use several well-known classification benchmarks for online learning, where the number of instances ranges from $1000$ to $581,012$.
All the experiments are performed over 20 different random permutations of the datasets. 
Besides, we introduce a large-scale real-world dataset \texttt{KuaiRec}~\cite{gao2022kuairec}, which has $4,494,578$ instances and associated timestamps. 
We do not tune the stepsizes $\eta$ of all the gradient descent-based algorithms but take the value $\eta = 0.2$.
We uniformly set $d=1$, $\alpha = 0.01$, $\eta_i=0$, $\sigma_i=0.5$ and $\lambda = 0.01$ for FORKS and SkeGD.
We take the Gaussian kernel $\kappa(\bm{x}_i,\bm{x}_j) = \exp\left(\frac{-||\bm{x}_i-\bm{x}_j||^2_2}{2\sigma^2}\right)$ with parameter set $\sigma\in\{2^{[-5:+0.5:7]}\}$ used by~\cite{zhang2019incremental}.
All experiments are performed on a machine with 24-core Intel(R) Xeon(R) Gold 6240R 2.40GHz CPU and 256 GB memory.

\subsection{More about KuaiRec Dataset}
\label{Apx: KuaiRec}
For our experiment, we utilize KuaiRec's small matrix as the dataset. 
The processing of dependent variables involves dividing the ratio of the user's time spent on the video to the video duration (watch ratio) by a threshold of 0.75, with values greater than 0.75 classified as positive and values less than or equal to 0.75 classified as negative. 
The selection of independent variables is obtained from three csv files, as specified in the code.

\subsection{Additional Experiment Results under Adversarial Environment}
\label{sec: Additional Experiment Results under Adversarial Environment}
We set $b=500$, $r=20$ for \texttt{codrna-2} and \texttt{german-2}.
The results are presented in Table~\ref{tbl: Ad2}. 

\begin{table}[!thbp]
\caption{Comparisons among BOGD, NOGD, PROS-N-KONS, SkeGD and our FORKS w.r.t. the mistake rates (\%) and the running time (s). The best result is highlighted in $\textbf{bold}$ font.}
\label{tbl: Ad2}
\footnotesize
\centering
\begin{spacing}{.93}
\resizebox{1.0\linewidth}{!}{
\begin{tabular}{lcccc}
\toprule
     \multirow{2}{*}{Algorithm}
             & \multicolumn{2}{c}{\texttt{codrna-2}}   & \multicolumn{2}{c}{\texttt{german-2}}
             \\\cmidrule(lr){2-3} \cmidrule(lr){4-5}
 & Mistake rate  &  Time  & Mistake rate  &  Time  \\\midrule
BOGD           &14.745 	$\pm$ 0.063 	&  0.043 	 &21.290 	$\pm$ 0.918 	&  0.060    \\
NOGD           &19.977 	$\pm$ 1.536 	&  0.041 	 &16.527 	$\pm$ 0.810 	&  0.056  \\
PROS-N-KONS    &15.430 	$\pm$ 2.315 	&  20.612 	 &11.187 	$\pm$ 1.782 	&  1.144     \\
SkeGD          &15.829 	$\pm$ 2.583 	&  0.203 	 &\ \ 5.742 	$\pm$ 2.647 	&  0.077  \\
FORKS       &\ \ \textbf{4.127} 	$\pm$ \textbf{0.769} 	&  0.039	 &\ \ \textbf{2.960} 	$\pm$ \textbf{0.185} 	&  0.050   \\
\bottomrule
\end{tabular}}
\end{spacing}
\end{table}

We demonstrate that all methods exhibit improved performance in a less hostile adversarial setting. 
Nevertheless, FORKS remains superior to other algorithms with significant advantages in terms of both time and prediction performance.

\subsection{Additional Experiment Results under Large-Scale Real-world Datasets}
\label{sec: Additional Experiment Results under Large-Scale Real-world Datasets}
Similar to the experimental setup described in section 5.3, we evaluate the performance of the algorithms across various budgets $B$, spanning from $100$ to $500$.
To avoid excessive training time, we use a budgeted version of PROS-N-KONS with a maximum budget of $B_{\max}=100$. 
Since the buffer size of PROS-N-KONS is data-dependent, we repeat the training process $20$ times to compute the average error rate and the logarithm of the average time for comparison.

\begin{figure}[!htp]
	\centering
	\begin{subfigure}{0.47\linewidth}
		\centering
		\includegraphics[width=\linewidth]{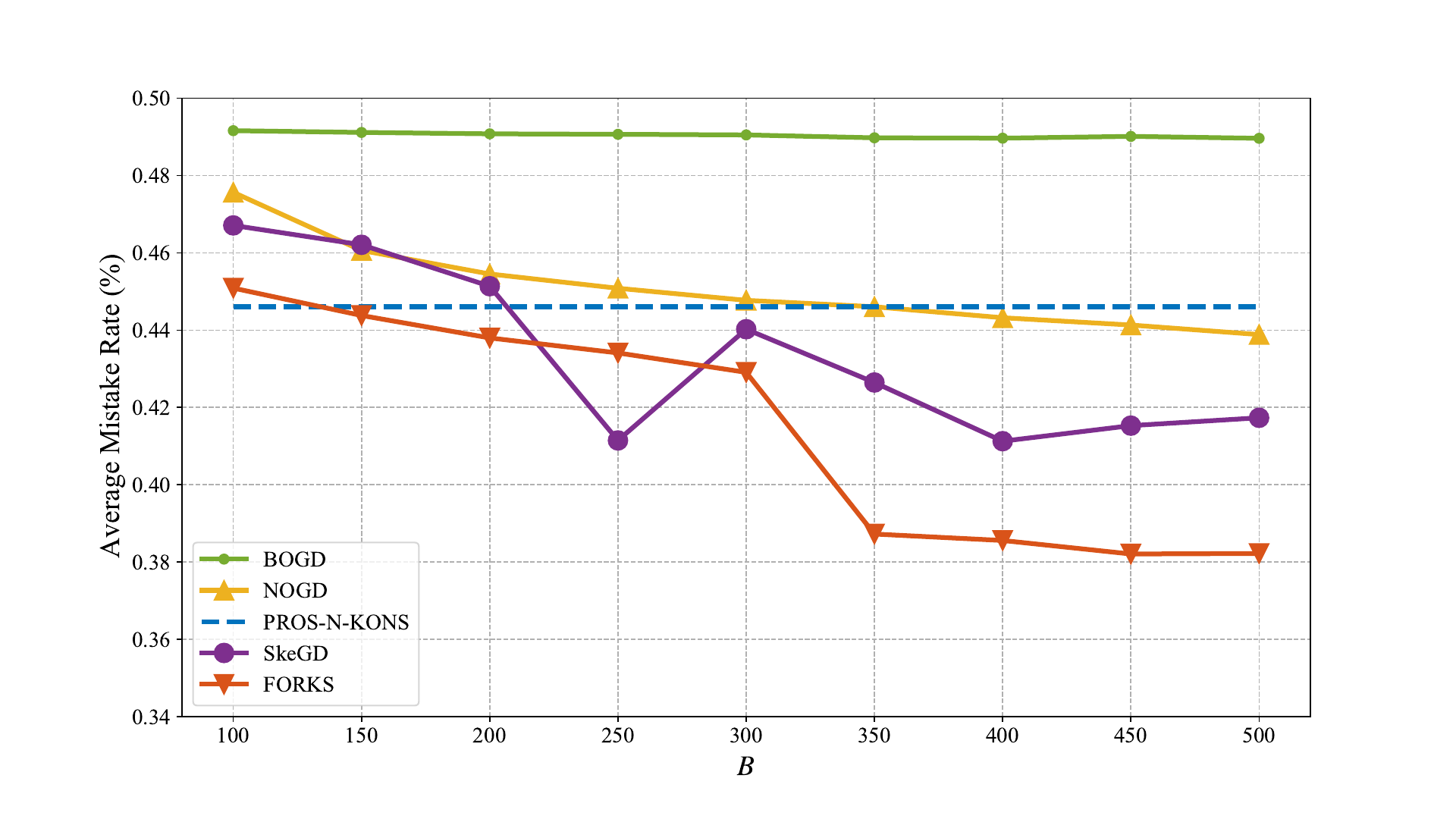}
		\caption{Average mistake rate}
		\label{Kuai_error_B}
	\end{subfigure}
	\centering
	\begin{subfigure}{0.47\linewidth}
		\centering
		\includegraphics[width=\linewidth]{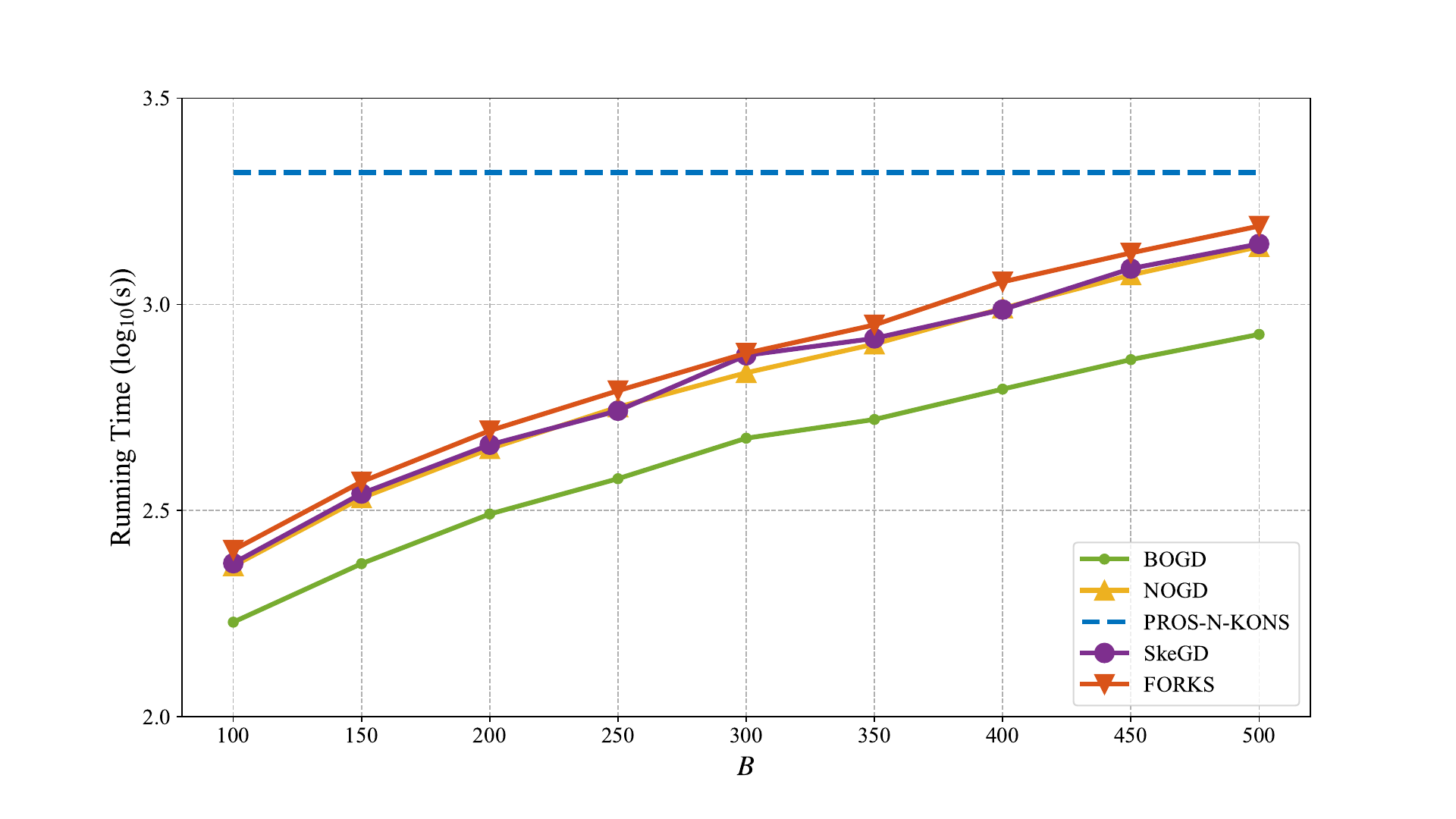}
		\caption{Average running time}
		\label{Kuai_time_B}
	\end{subfigure}
	\caption{The mistake rates and average running time on \texttt{KuaiRec} under hinge loss. As PROS-N-KONS utilizes an adaptive budget, it cannot modify computational costs, thereby being depicted as a parallel line in the figures.}
    \label{Kuai_exp_B}
\end{figure}

From Figure~\ref{Kuai_exp_B}, we can observe that our FORKS show the best learning performance under most budget conditions. 
The large-scale experiments validate the effectiveness and efficiency of our proposed FORKS, making it potentially more practical than the existing second-order online kernel learning approaches. 
Meanwhile, we observed that increasing the budget size $B$ results in a lower mistake rate but also leads to a higher computation time cost. 
In practice, we can flexibly adjust the budget size based on our estimates of the data stream size to obtain a better approximation quality.

\newpage

\bibliographystyle{named}
\bibliography{ijcai25}

\begin{thebibliography}{}

\bibitem[\protect\citeauthoryear{Belkin}{2018}]{belkin2018approximation}
Mikhail Belkin.
\newblock Approximation beats concentration? an approximation view on inference with smooth radial kernels.
\newblock In {\em {COLT}}, volume~75, pages 1348--1361, 2018.

\bibitem[\protect\citeauthoryear{Brand}{2006}]{brand2006fast}
Matthew Brand.
\newblock Fast low-rank modifications of the thin singular value decomposition.
\newblock {\em Linear algebra and its applications}, 415(1):20--30, 2006.

\bibitem[\protect\citeauthoryear{Calandriello \bgroup \em et al.\egroup }{2017a}]{calandriello2017efficient}
Daniele Calandriello, Alessandro Lazaric, and Michal Valko.
\newblock Efficient second-order online kernel learning with adaptive embedding.
\newblock In {\em {NIPS}}, pages 6140--6150, 2017.

\bibitem[\protect\citeauthoryear{Calandriello \bgroup \em et al.\egroup }{2017b}]{calandriello2017second}
Daniele Calandriello, Alessandro Lazaric, and Michal Valko.
\newblock Second-order kernel online convex optimization with adaptive sketching.
\newblock In {\em {ICML}}, volume~70, pages 645--653, 2017.

\bibitem[\protect\citeauthoryear{CAO \bgroup \em et al.\egroup }{2017}]{LeleCAO:276}
Lele CAO, Fuchun SUN, Hongbo LI, and Wenbing HUANG.
\newblock Advancing the incremental fusion of robotic sensory features using online multi-kernel extreme learning machine.
\newblock {\em Frontiers of Computer Science}, 11:276, 2017.

\bibitem[\protect\citeauthoryear{Cavallanti \bgroup \em et al.\egroup }{2007}]{cavallanti2007tracking}
Giovanni Cavallanti, Nicol{\`{o}} Cesa{-}Bianchi, and Claudio Gentile.
\newblock Tracking the best hyperplane with a simple budget perceptron.
\newblock {\em Mach. Learn.}, 69(2-3):143--167, 2007.

\bibitem[\protect\citeauthoryear{Charikar \bgroup \em et al.\egroup }{2002}]{charikar2002finding}
Moses Charikar, Kevin~C. Chen, and Martin Farach{-}Colton.
\newblock Finding frequent items in data streams.
\newblock In {\em {ICALP}}, volume 2380, pages 693--703, 2002.

\bibitem[\protect\citeauthoryear{CHEN \bgroup \em et al.\egroup }{2022}]{ChengCHEN:165330}
Cheng CHEN, Weinan ZHANG, and Yong YU.
\newblock Efficient policy evaluation by matrix sketching.
\newblock {\em Frontiers of Computer Science}, 16:165330, 2022.

\bibitem[\protect\citeauthoryear{Dekel \bgroup \em et al.\egroup }{2005}]{DekelSS05}
Ofer Dekel, Shai Shalev{-}Shwartz, and Yoram Singer.
\newblock The forgetron: {A} kernel-based perceptron on a fixed budget.
\newblock In {\em {NIPS}}, pages 259--266, 2005.

\bibitem[\protect\citeauthoryear{Gao \bgroup \em et al.\egroup }{2022}]{gao2022kuairec}
Chongming Gao, Shijun Li, Wenqiang Lei, Jiawei Chen, Biao Li, Peng Jiang, Xiangnan He, Jiaxin Mao, and Tat{-}Seng Chua.
\newblock Kuairec: {A} fully-observed dataset and insights for evaluating recommender systems.
\newblock In {\em {CIKM}}, pages 540--550, 2022.

\bibitem[\protect\citeauthoryear{Hazan \bgroup \em et al.\egroup }{2007}]{hazan2007logarithmic}
Elad Hazan, Amit Agarwal, and Satyen Kale.
\newblock Logarithmic regret algorithms for online convex optimization.
\newblock {\em Mach. Learn.}, 69(2-3):169--192, 2007.

\bibitem[\protect\citeauthoryear{Hoi \bgroup \em et al.\egroup }{2012}]{zhao2012fast}
Steven C.~H. Hoi, Jialei Wang, Peilin Zhao, Rong Jin, and Pengcheng Wu.
\newblock Fast bounded online gradient descent algorithms for scalable kernel-based online learning.
\newblock In {\em {ICML}}, 2012.

\bibitem[\protect\citeauthoryear{Hu \bgroup \em et al.\egroup }{2015}]{Hu2015Kernelized}
Junjie Hu, Haiqin Yang, Irwin King, Michael~R. Lyu, and Anthony~Man{-}Cho So.
\newblock Kernelized online imbalanced learning with fixed budgets.
\newblock In {\em {AAAI}}, pages 2666--2672, 2015.

\bibitem[\protect\citeauthoryear{Kane and Nelson}{2014}]{kane2014sparser}
Daniel~M. Kane and Jelani Nelson.
\newblock Sparser johnson-lindenstrauss transforms.
\newblock {\em J. {ACM}}, 61(1):4:1--4:23, 2014.

\bibitem[\protect\citeauthoryear{Kivinen \bgroup \em et al.\egroup }{2001}]{kivinen2004online}
Jyrki Kivinen, Alexander~J. Smola, and Robert~C. Williamson.
\newblock Online learning with kernels.
\newblock In {\em {NIPS}}, pages 785--792, 2001.

\bibitem[\protect\citeauthoryear{Liu and Liao}{2015}]{liu2015eigenvalues}
Yong Liu and Shizhong Liao.
\newblock Eigenvalues ratio for kernel selection of kernel methods.
\newblock In {\em {AAAI}}, pages 2814--2820, 2015.

\bibitem[\protect\citeauthoryear{Lu \bgroup \em et al.\egroup }{2016a}]{lu2016large}
Jing Lu, Steven C.~H. Hoi, Jialei Wang, Peilin Zhao, and Zhiyong Liu.
\newblock Large scale online kernel learning.
\newblock {\em J. Mach. Learn. Res.}, 17:47:1--47:43, 2016.

\bibitem[\protect\citeauthoryear{Lu \bgroup \em et al.\egroup }{2016b}]{Lu2016Online}
Jing Lu, Peilin Zhao, and Steven C.~H. Hoi.
\newblock Online sparse passive aggressive learning with kernels.
\newblock In {\em {SDM}}, pages 675--683, 2016.

\bibitem[\protect\citeauthoryear{Luo \bgroup \em et al.\egroup }{2016}]{luo2016efficient}
Haipeng Luo, Alekh Agarwal, Nicol{\`{o}} Cesa{-}Bianchi, and John Langford.
\newblock Efficient second order online learning by sketching.
\newblock In {\em {NIPS}}, pages 902--910, 2016.

\bibitem[\protect\citeauthoryear{Orabona \bgroup \em et al.\egroup }{2008}]{orabona2008projectron}
Francesco Orabona, Joseph Keshet, and Barbara Caputo.
\newblock The projectron: a bounded kernel-based perceptron.
\newblock In {\em {ICML}}, volume 307, pages 720--727, 2008.

\bibitem[\protect\citeauthoryear{Sahoo \bgroup \em et al.\egroup }{2019}]{Sahoo2019Large}
Doyen Sahoo, Steven C.~H. Hoi, and Bin Li.
\newblock Large scale online multiple kernel regression with application to time-series prediction.
\newblock {\em {ACM} Trans. Knowl. Discov. Data}, 13(1):9:1--9:33, 2019.

\bibitem[\protect\citeauthoryear{Shalev{-}Shwartz}{2012}]{Shalev2011OLA}
Shai Shalev{-}Shwartz.
\newblock Online learning and online convex optimization.
\newblock {\em Found. Trends Mach. Learn.}, 4(2):107--194, 2012.

\bibitem[\protect\citeauthoryear{Singh \bgroup \em et al.\egroup }{2012}]{Singh2012Online}
Abhishek Singh, Narendra Ahuja, and Pierre Moulin.
\newblock Online learning with kernels: Overcoming the growing sum problem.
\newblock In {\em {MLSP}}, pages 1--6, 2012.

\bibitem[\protect\citeauthoryear{Wan and Zhang}{2022}]{wan2021efficient}
Yuanyu Wan and Lijun Zhang.
\newblock Efficient adaptive online learning via frequent directions.
\newblock {\em {IEEE} Trans. Pattern Anal. Mach. Intell.}, 44(10):6910--6923, 2022.

\bibitem[\protect\citeauthoryear{Wang and Vucetic}{2010}]{wang2010online}
Zhuang Wang and Slobodan Vucetic.
\newblock Online passive-aggressive algorithms on a budget.
\newblock In {\em {AISTATS}}, volume~9, pages 908--915, 2010.

\bibitem[\protect\citeauthoryear{Wang \bgroup \em et al.\egroup }{2016}]{wang2016spsd}
Shusen Wang, Luo Luo, and Zhihua Zhang.
\newblock {SPSD} matrix approximation vis column selection: Theories, algorithms, and extensions.
\newblock {\em J. Mach. Learn. Res.}, 17:49:1--49:49, 2016.

\bibitem[\protect\citeauthoryear{Wang \bgroup \em et al.\egroup }{2018}]{wang2018minimizing}
Guanghui Wang, Dakuan Zhao, and Lijun Zhang.
\newblock Minimizing adaptive regret with one gradient per iteration.
\newblock In {\em {IJCAI}}, pages 2762--2768, 2018.

\bibitem[\protect\citeauthoryear{Williams and Seeger}{2000}]{williams2000using}
Christopher K.~I. Williams and Matthias~W. Seeger.
\newblock Using the nystr{\"{o}}m method to speed up kernel machines.
\newblock In {\em {NIPS}}, pages 682--688, 2000.

\bibitem[\protect\citeauthoryear{Woodruff}{2014}]{Woodruff2014SAA}
David~P. Woodruff.
\newblock Sketching as a tool for numerical linear algebra.
\newblock {\em Found. Trends Theor. Comput. Sci.}, 10(1-2):1--157, 2014.

\bibitem[\protect\citeauthoryear{Yang \bgroup \em et al.\egroup }{2012}]{yang2012nystrom}
Tianbao Yang, Yufeng Li, Mehrdad Mahdavi, Rong Jin, and Zhi{-}Hua Zhou.
\newblock Nystr{\"{o}}m method vs random fourier features: {A} theoretical and empirical comparison.
\newblock In {\em {NIPS}}, pages 485--493, 2012.

\bibitem[\protect\citeauthoryear{Zhang and Liao}{2018}]{Zhang2018Online}
Xiao Zhang and Shizhong Liao.
\newblock Online kernel selection via incremental sketched kernel alignment.
\newblock In {\em IJCAI}, pages 3118--3124, 2018.

\bibitem[\protect\citeauthoryear{Zhang and Liao}{2019}]{zhang2019incremental}
Xiao Zhang and Shizhong Liao.
\newblock Incremental randomized sketching for online kernel learning.
\newblock In {\em {ICML}}, volume~97, pages 7394--7403, 2019.

\bibitem[\protect\citeauthoryear{Zhang \bgroup \em et al.\egroup }{2023}]{zhang2023reward}
Xiao Zhang, Ninglu Shao, Zihua Si, Jun Xu, Wenhan Wang, Hanjing Su, and Ji-Rong Wen.
\newblock Reward imputation with sketching for contextual batched bandits.
\newblock {\em NIPS}, pages 64577--64588, 2023.

\bibitem[\protect\citeauthoryear{Zhdanov and Kalnishkan}{2013}]{zhdanov2013identity}
Fedor Zhdanov and Yuri Kalnishkan.
\newblock An identity for kernel ridge regression.
\newblock {\em Theor. Comput. Sci.}, 473:157--178, 2013.

\bibitem[\protect\citeauthoryear{Zhu and Xu}{2015}]{zhu2015online}
Changbo Zhu and Huan Xu.
\newblock Online gradient descent in function space.
\newblock {\em arXiv}, 2015.

\bibitem[\protect\citeauthoryear{Zinkevich}{2003}]{zinkevich2003online}
Martin Zinkevich.
\newblock Online convex programming and generalized infinitesimal gradient ascent.
\newblock In {\em {ICML}}, pages 928--936, 2003.

\end{thebibliography}

\end{document}